\documentclass[11pt,a4paper]{article}
\usepackage{times,latexsym}
\usepackage{url}
\usepackage[T1]{fontenc}

\usepackage{soulutf8}
\usepackage{graphicx}
\usepackage{amsmath}
\usepackage{booktabs}
\usepackage{tabularx}
\usepackage{array,multirow}
\usepackage{subcaption}
\usepackage{float}
\usepackage{makecell}
\usepackage{framed}
\usepackage{enumitem}
\usepackage{titlesec} 
\usepackage{microtype}
\usepackage{siunitx}
\usepackage{anyfontsize}
\usepackage{xcolor}

\setlist{nosep,topsep=3pt}

\newcommand{\STAB}[1]{\begin{tabular}{@{}c@{}}#1\end{tabular}}
\newcommand{\rotaterow}[2]{\multirow{#1}{*}{\STAB{\rotatebox[origin=c]{90}{#2}}}}
\newcommand{\multiline}[1]{\begin{tabular}{@{}c@{}}#1\end{tabular}}

\definecolor{lexcolor}{rgb}{.1, .1, .4}
\newcommand{\lex}[1]{{\textit{#1}}}

\newcommand{\qargu}[3]{\begin{quote}\textbf{Claim:} \lex{#1} \\[1mm] \textbf{#2 Statement:} \lex{#3}\end{quote}}
\newcommand{\qqargu}[5]{\begin{quote}\textbf{Claim:} \lex{#1} \\[1mm] \textbf{#2 Statement:} \lex{#3} \\[1mm] \textbf{#4 Statement:} \lex{#5}\end{quote}}

\newcommand{\hide}[1]{}

\newcolumntype{C}{>{\centering\arraybackslash}X}

\definecolor{hlcolor}{rgb}{.9, 1, .4}
\sethlcolor{hlcolor}

\colorlet{shadecolor}{hlcolor}

\newcommand{\pred}[2]{\text{#1}(#2)}
\newcommand{\pslrule}[2]{$#1 \rightarrow #2$}

\newcommand{\fontsizeto}[1]{\fontsize{#1}{#1}\selectfont}
\newenvironment{resize}[1]{\par\vspace{-10pt}\fontsizeto{#1}}{\hspace{-3pt}}

\newcommand{\Fsup}{F1$_{\text{sup}}$}
\newcommand{\Fatt}{F1$_{\text{att}}$}
\newcommand{\Fneu}{F1$_{\text{neu}}$}

\renewenvironment{quote}
  {\list{}{\leftmargin=.5cm \rightmargin=.5cm \topsep=2pt}\item\relax}
  {\endlist}
\expandafter\def\expandafter\quote\expandafter{\quote\fontsize{10}{11.5}\selectfont}

\usepackage[acceptedWithA]{tacl2018v2}

\usepackage{xspace,mfirstuc,tabulary}

\newif\iftaclinstructions
\taclinstructionsfalse 
\iftaclinstructions
\renewcommand{\confidential}{}
\renewcommand{\anonsubtext}{(No author info supplied here, for consistency with
TACL-submission anonymization requirements)}
\newcommand{\instr}
\fi

\iftaclpubformat 

\else

\fi

\title{Classifying Argumentative Relations\\Using Logical Mechanisms and Argumentation Schemes}

\author{
 Yohan Jo$^1$ ~~ Seojin Bang$^1$ ~~ Chris Reed$^2$ ~~ Eduard Hovy$^1$ \\
 $^1$School of Computer Science, Carnegie Mellon University, USA \\
 $^2$Centre for Argument Technology, University of Dundee, UK  \\
  $^{1}${\sf \{yohanj,seojinb,ehovy\}@andrew.cmu.edu}, $^2${\sf c.a.reed@dundee.ac.kr} \\
}

\date{}

\begin{document}
\maketitle
\begin{abstract}
While argument mining has achieved significant success in classifying argumentative relations between statements (support, attack, and neutral), we have a limited computational understanding of logical mechanisms that constitute those relations. 
Most recent studies rely on black-box models, which are not as linguistically insightful as desired. On the other hand, earlier studies use rather simple lexical features, missing logical relations between statements.
To overcome these limitations, our work classifies argumentative relations based on four logical and theory-informed mechanisms between two statements, namely (i) factual consistency, (ii) sentiment coherence, (iii) causal relation, and (iv) normative relation. 
We demonstrate that our operationalization of these logical mechanisms classifies argumentative relations without directly training on data labeled with the relations, significantly better than several unsupervised baselines. 
We further demonstrate that these mechanisms also improve supervised classifiers through representation learning.
\end{abstract}

\section{Introduction}
There have been great advances in argument mining---classifying the argumentative relation between statements as support, attack, or neutral. Recent research has focused on training complex neural networks on large labeled data. However, the behavior of such models remains obscure, and recent studies found evidence that those models may rely on spurious statistics of training data~\cite{Niven:2019cq} and superficial cues irrelevant to the meaning of statements, such as discourse markers~\cite{Opitz:2019tk}. Hence, in this work, we turn to an \textit{interpretable} method to investigate \textit{logical relations} between statements, such as causal relations and factual contradiction.
Such relations have been underemphasized in earlier studies~\cite{Feng:2011vp,Lawrence:2016kt}, possibly because their operationalization was unreliable then.
Now that computational semantics is fast developing, our work takes a first step to computationally investigate how logical mechanisms contribute to building argumentative relations between statements and to classification accuracy with and without training on labeled data. 

To investigate what logical mechanisms govern argumentative relations, we hypothesize that governing mechanisms should be able to classify the relations without directly training on relation-labeled data. Thus, we first compile a set of rules specifying logical and theory-informed mechanisms that signal the support and attack relations (\S{\ref{sec:rules}}). The rules are grouped into four mechanisms: factual consistency, sentiment coherence, causal relation, and normative relation. These rules are combined via probabilistic soft logic (PSL)~\cite{Bach:2017psl} to estimate the optimal argumentative relations between statements. 
We operationalize each mechanism by training semantic modules on public datasets so that the modules reflect real-world knowledge necessary for reasoning (\S{\ref{sec:modules}}). 
For normative relation, we build a necessary dataset
via rich annotation of the normative argumentation schemes \textit{argument from consequences} and \textit{practical reasoning} \cite{Walton:2008schem}, by developing a novel and reliable annotation protocol (\S{\ref{sec:annotation}}). 

Our evaluation is based on arguments from kialo.com and debatepedia.org. 
We first demonstrate that the four logical mechanisms explain the argumentative relations between statements effectively. PSL with our operationalization of the mechanisms can classify the relations without direct training on relation-labeled data, outperforming several unsupervised baselines (\S{\ref{sec:exp1}}). We analyze the contribution and pitfalls of individual mechanisms in detail.
Next, to examine whether the mechanisms can further inform supervised models, we present a method to learn vector representations of arguments that are ``cognizant of'' the logical mechanisms (\S{\ref{sec:exp2}}). This method outperforms several supervised models trained without concerning the mechanisms, as well as models that incorporate the mechanisms in different ways. We illustrate how it makes a connection between logical mechanisms and argumentative relations. 
Our contributions are:
\begin{itemize}
    \item An interpretable method based on PSL to investigate logical and theory-informed mechanisms in argumentation computationally. 
    \item A representation learning method that incorporates the logical mechanisms to improve the predictive power of supervised models. 
    \item A novel and reliable annotation protocol, along with a rich schema, for the argumentation schemes \textit{argument from consequences} and \textit{practical reasoning}. We release our annotation manuals and annotated data.\footnote{The annotations, data, and source code are available at: \url{https://github.com/yohanjo/tacl_arg_rel}}
\end{itemize}

\section{Related Work\label{sec:related_work}}
There has been active research in NLP to understand different mechanisms of argumentation computationally. 
Argumentative relations have been found to be associated with various statistics, such as discourse markers~\cite{Opitz:2019tk}, sentiment~\cite{Allaway:2020stance}, and use of negating words~\cite{Niven:2019cq}. Further, as framing plays an important role in debates~\cite{Ajjour:2019frame}, different stances for a topic emphasize different points, resulting in strong thematic correlations~\cite{Lawrence:2017topic}. 

Such thematic associations have been exploited in stance detection and dis/agreement classification. Stance detection~\cite{Allaway:2020stance,Stab:2018cross-topic,Xu.2018} aims to classify a statement as pro or con with respect to a topic, while dis/agreement classification~\cite{Chen:2018hybrid,Hou.2017,Rosenthal:2015iw} aims to decide whether two statements are from the same or opposite stance(s) for a given topic. Topics are usually discrete, and models often learn thematic correlations between a topic and a stance~\cite{Xu:2019reason}. Our work is slightly different as we classify the \textit{direct} support or attack relation between two \textit{natural} statements.

The aforementioned correlations, however, are rather byproducts than core mechanisms of argumentative relations. In order to decide whether a statement supports or attacks another, we cannot ignore the \textit{logical} relation between them. Textual entailment was found to inform argumentative relations~\cite{choi-lee-2018-gist} and used to detect arguments~\cite{Cabrio:2012ud}. Similarly, there is evidence that the opinions of two statements toward the same concept constitute their argumentative relations~\cite{Gemechu2020:acl,Kobbe.2020}. Causality between events also received attention, and causality graph construction was proposed for argument analysis~\cite{Al-Khatib:2020.causal_graph}.
Additionally, in argumentation theory, Walton's argumentation schemes~\cite{Walton:2008schem} specify common reasoning patterns people use to form an argument. This motivates our work to investigate logical mechanisms in four categories: factual consistency, sentiment coherence, causal relation, and normative relation. 

Logical mechanisms have not been actively studied in argumentative relation classification. Models based on hand-crafted features have used relatively simple lexical features, such as $n$-grams, discourse markers, and sentiment agreement and word overlap between two statements~\cite{Stab:2017cl,Habernal:2017cl,Persing:2016e2e,Rinott.2015.evidence}. 
Recently, neural models have become dominant approaches~\cite{Chakrabarty:2019ampersand,Durmus:2019kialo,Eger:2017e2e}. Despite their high accuracy and finding of some word-level interactions between statements~\cite{Xu:2019reason,Chen:2018hybrid}, they provide quite limited insight into governing mechanisms in argumentative relations.
Indeed, more and more evidence suggests that supervised models learn to overly rely on superficial cues, such as discourse markers~\cite{Opitz:2019tk}, negating words~\cite{Niven:2019cq}, and sentiment~\cite{Allaway:2020stance} behind the scenes. 
We instead use an interpretable method based on PSL to examine logical mechanisms (\S{\ref{sec:exp1}}) and then show evidence that these mechanisms can inform supervised models in intuitive ways (\S{\ref{sec:exp2}}).

Some research adopted argumentation schemes as a framework, making comparisons with discourse relations~\cite{Cabrio.2013.schemes} and collecting and leveraging data at varying degrees of granularity. At a coarse level, prior studies annotated the presence of particular argumentation schemes in text~\cite{Visser.2020.Argumentation,Lawrence:2019we,Lindahl:2019uc,Reed.2008.araucaria} and developed models to classify different schemes~\cite{Feng:2011vp}. However, each scheme often accommodates both support and attack relations between statements, so classifying those relations requires semantically richer information within the scheme than just its presence. To that end, \citet{Reisert.2018.schemes} annotated individual components within schemes, particularly emphasizing \textit{argument from consequences}. Based on the logic behind this scheme, \citet{Kobbe.2020} developed an unsupervised method to classify the support and attack relations using syntactic rules and lexicons. Our work extends these studies by including other normative schemes (\textit{practical reasoning} and property-based reasoning) and annotating richer information.

\section{Rules\label{sec:rules}}

We first compile rules that specify evidence for the support and attack relations between \textbf{claim} $\boldsymbol{C}$ and \textbf{statement} $\boldsymbol{S}$ (Table~\ref{tab:psl_rules})\footnote{We do not assume that claim-hood and statement-hood are intrinsic features of text spans; we follow prevailing argumentation theory in viewing claims and statements as roles determined by virtue of relationships between text spans.}. 
These rules are combined via probabilistic soft logic (PSL) \cite{Bach:2017psl} to estimate the optimal relation between $C$ and $S$\footnote{Predicates in the rules are probability scores, and PSL aims to estimate the scores of Support($S,C$), Attack($S,C$), and Neutral($S,C$) for all ($S,C$). The degree of satisfaction of the rules are converted to a loss, which is minimized via maximum likelihood estimation.}. 

We will describe individual rules in four categories: factual consistency, sentiment coherence, causal relation, and normative relation, followed by additional chain rules.


\subsection{Factual Consistency}
A statement that supports the claim may present a fact that naturally entails the claim, while an attacking statement often presents a fact contradictory or contrary to the claim. For example:
\qqargu{Homeschooling deprives children and families from interacting with people with different religions, ideologies or values.}{Support}{Homeschool students have few opportunities to meet diverse peers they could otherwise see at normal schools.}{Attack}{Homeschool students can interact regularly with other children from a greater diversity of physical locations, allowing them more exposure outside of their socio-economic group.}
This logic leads to two rules:
\begin{resize}{10}
\begin{align*}
    &\textbf{R1: } &&\pred{FactEntail}{S,C} \rightarrow \pred{Support}{S,C}, \\
    &\textbf{R2: } &&\pred{FactContradict}{S,C} \rightarrow \pred{Attack}{S,C}
\end{align*}\vspace{-8mm}
\begin{align*}
    \textit{s.t. } &\pred{FactEntail}{S,C} = P(S \text{ entails } C), \\
    &\pred{FactContradict}{S,C} = P(S \text{ contradicts } C).
\end{align*}
\end{resize}
In our work, these probabilities are computed by a textual entailment module (\S{\ref{sec:mod_text_entail}}).

In argumentation, it is often the case that an attacking statement and the claim are not strictly contradictory nor contrary, but the statement contradicts only a specific part of the claim, as in:
\qargu{\ul{Vegan diets} are healthy.}{Attack}{\ul{Meat} is healthy.}
Formally, let $(A^S_{i,0}, A^S_{i,1}, \cdots)$ denote the $i$th relation tuple in $S$, and $(A^C_{j,0}, A^C_{j,1}, \cdots)$ the $j$th relation tuple in $C$. We formulate the conflict rule:
\begin{resize}{10}
\begin{align*}
    &\textbf{R3: } \pred{FactConflict}{S,C} \rightarrow \pred{Attack}{S,C}
\end{align*}
\end{resize}\vspace{-8mm}
\begin{resize}{9}
\begin{align*}
    &\textit{s.t. } \pred{FactConflict}{S,C} = \\
    &~ \max_{i,j,k} P(A^S_{i,k} \text{ contradicts } A^C_{j,k})\prod_{k' \neq k} P(A^S_{i,k'} \text{ entails } A^C_{j,k'}).
\end{align*}
\end{resize}
We use Open IE 5.1\hide{\footnote{\url{https://git.io/JTr3Y}}} to extract relation tuples, and the probability terms are computed by a textual entailment module (\S{\ref{sec:mod_text_entail}}).

\begin{table}[t]
    \footnotesize
    \centering
    \begin{tabularx}{\linewidth}{p{3mm}p{.2cm}X} \toprule
         & & Rules  \\
        \midrule
        \rotaterow{3}{\shortstack{Factual\\Consist.}} 
         & R1 & \pslrule{\pred{FactEntail}{S,C}}{\pred{Support}{S,C}} \\
         & R2 & \pslrule{\pred{FactContradict}{S,C}}{\pred{Attack}{S,C}}  \\
         & R3 & \pslrule{\pred{FactConflict}{S,C}}{\pred{Attack}{S,C}}  \\
        \midrule
        \rotaterow{2}{\shortstack{Senti\\Cohe.}} 
         & R4 & \pslrule{\pred{SentiConflict}{S,C}}{\pred{Attack}{S,C}}  \\
         & R5 & \pslrule{\pred{SentiCoherent}{S,C}}{\pred{Support}{S,C}}  \\
        \midrule
        \rotaterow{6}{\shortstack{Causal\\Relation}} & 
         & \textbf{\textit{CAUSE-TO-EFFECT ~REASONING}} \\
         & R6 & \pslrule{\pred{Cause}{S,C}}{\pred{Support}{S,C}} \\
         & R7 & \pslrule{\pred{Obstruct}{S,C}}{\pred{Attack}{S,C}} \\ 
        \cmidrule(lr){2-3}
         & & \textbf{\textit{EFFECT-TO-CAUSE ~REASONING}} \\
         & R8 & \pslrule{\pred{Cause}{C,S}}{\pred{Support}{S,C}}  \\
         & R9 & \pslrule{\pred{Obstruct}{C,S}}{\pred{Attack}{S,C}}  \\ 
        \midrule
        \rotaterow{6}{\shortstack{Normative\\Relation}} & 
         & \textbf{\textit{ARGUMENT ~FROM ~CONSEQUENCES}} \\
         & R10 & \pslrule{\pred{BackingConseq}{S,C}}{\pred{Support}{S,C}}  \\
         & R11 & \pslrule{\pred{RefutingConseq}{S,C}}{\pred{Attack}{S,C}}  \\
        \cmidrule(lr){2-3}        
         & & \textbf{\textit{PRACTICAL ~REASONING}}  \\
         & R12 & \pslrule{\pred{BackingNorm}{S,C}}{\pred{Support}{S,C}} \\
         & R13 & \pslrule{\pred{RefutingNorm}{S,C}}{\pred{Attack}{S,C}}  \\ 
        \midrule
        \rotaterow{4}{\shortstack{Relation\\Chain}}
         & R14 & \shortstack{\pslrule{\pred{Support}{S,I} \wedge \pred{Support}{I,C}}{\pred{Support}{S,C}}} \\
         & R15 & \shortstack{\pslrule{\pred{Attack}{S,I} \wedge \pred{Attack}{I,C}}{\pred{Support}{S,C}}} \\
         & R16 & \shortstack{\pslrule{\pred{Support}{S,I} \wedge \pred{Attack}{I,C}}{\pred{Attack}{S,C}}}  \\
         & R17 & \shortstack{\pslrule{\pred{Attack}{S,I} \wedge \pred{Support}{I,C}}{\pred{Attack}{S,C}}}  \\
        \midrule
        \rotaterow{2}{\shortstack{Const-\\raints}}
         & C1 & $\pred{Neutral}{S,C}$ = 1 \\
         & C2 & \shortstack{$\pred{Support}{S,C}$+$\pred{Attack}{S,C}$+$\pred{Neutral}{S,C}$ = 1} \\
        \bottomrule
    \end{tabularx}
    \caption{PSL rules. ($S$: statement, $C$: claim)}
    \label{tab:psl_rules}
\end{table}

\subsection{Sentiment Coherence}
When $S$ attacks $C$, they may express opposite sentiments toward the same target, whereas they may express the same sentiment if $S$ supports $C$~\cite{Gemechu2020:acl}. For example:
\qqargu{Pet keeping is morally justified.}{Attack}{Keeping pets is hazardous and offensive to other people.}{Support}{Pet owners can provide safe places and foods to pets.}

Let $(t^S_i, s^S_i)$ be the $i$th expression of sentiment $s^S_i \in \{\text{pos, neg, neu}\}$ toward target $t^S_i$ in $S$, and $(t^C_j, s^C_j)$ the $j$th expression in $C$. We formulate two rules:
\begin{resize}{10}
\begin{align*}
    &\textbf{R4:} &&\pred{SentiConflict}{S,C} \rightarrow \pred{Attack}{S,C}, \\
    &\textbf{R5:} &&\pred{SentiCoherent}{S,C} \rightarrow \pred{Support}{S,C}
\end{align*}\vspace{-6mm}
\begin{align*}
    \textit{s.t. } &\pred{SentiConflict}{S,C} = \\
    &~~~~~~ \max_{i,j} P(t^S_i = t^C_j) 
    \left\{ P(s^S_i = \text{pos}) P(s^C_j = \text{neg}) \right. \\
    &~~~~~~~~~~~~~~~~~~~~~~~ \left. + P(s^S_i = \text{neg}) P(s^C_j = \text{pos}) \right\}, \displaybreak[0] \\
    &\pred{SentiCoherent}{S,C} = \\
    &~~~~~~ \max_{i,j} P(t^S_i = t^C_j) 
    \left\{ P(s^S_i = \text{pos}) P(s^C_j = \text{pos}) \right. \\
    &~~~~~~~~~~~~~~~~~~~~~~~ \left. + P(s^S_i = \text{neg}) P(s^C_j = \text{neg}) \right\}.
\end{align*}
\end{resize}
In this work, targets are all noun phrases and verb phrases in $C$ and $S$. $P(t^S_i = t^C_j)$ is computed by a textual entailment module (\S{\ref{sec:mod_text_entail}}), and $P(s^{S}_{i})$ and $P(s^{C}_j)$ by a target-based sentiment classifier (\S{\ref{sec:mod_senti_analy}}).

\subsection{Causal Relation}
Reasoning based on causal relation between events is used in two types of argumentation: \textit{argument from cause to effect} and \textit{argument from effect to cause} \cite{Walton:2008schem}. 
In cause-to-effect (C2E) reasoning, $C$ is derived from $S$ because the event in $S$ may cause that in $C$. If $S$ causes (obstructs) $C$ then $S$ is likely to support (attack) $C$. For example:
\qqargu{Walmart's stock price will rise.}{Support}{Walmart generated record revenue.}{Attack}{Walmart had low net incomes.}
This logic leads to two rules:
\begin{resize}{10}
\begin{align*}
    &\textbf{R6:} &&\pred{Cause}{S,C} \rightarrow \pred{Support}{S,C}, \\
    &\textbf{R7:} &&\pred{Obstruct}{S,C} \rightarrow \pred{Attack}{S,C},
\end{align*}\vspace{-6mm}
\begin{align*}
    \textit{s.t. } &\pred{Cause}{S,C} = P(S \text{ causes } C), \\
    &\pred{Obstruct}{S,C} = P(S \text{ obstructs } C).
\end{align*}
\end{resize}

\vspace{-2mm}
Effect-to-cause (E2C) reasoning has the reversed direction; $S$ describes an observation and $C$ is a reasonable explanation that may have caused it. If $C$ causes (obstructs) $S$, then $S$ is likely to support (attack) $C$, as in:
\qqargu{St. Andrew Art Gallery is closing soon.}{Support}{The number of paintings in the gallery has reduced by half for the past month.}{Attack}{The gallery recently bought 20 photographs.}

\begin{resize}{10}
\begin{align*}
    &\textbf{R8:} &&\pred{Cause}{C,S} \rightarrow \pred{Support}{S,C}, \\
    &\textbf{R9:} &&\pred{Obstruct}{C,S} \rightarrow \pred{Attack}{S,C}.
\end{align*}
\end{resize}
The probabilities are computed by a causality module (\S{\ref{sec:mod_causality}}).

\subsection{Normative Relation\label{sec:rules_schemes}}
In argumentation theory, Walton's argumentation schemes specify common reasoning patterns used in arguments~\cite{Walton:2008schem}. 
We focus on two schemes related to normative arguments, whose claims suggest that an action or situation be brought about. Normative claims are one of the most common proposition types in argumentation~\cite{Jo:2020lrec} and have received much attention in the literature~\cite{Park:2018wy}.

\paragraph{Argument from Consequences:}
In this scheme, the claim is supported or attacked by a positive or negative consequence, as in:
\qqargu{Humans should stop eating animal meat.}{Support}{The normalizing of killing animals for food leads to a cruel mankind. \hfill\hfill\textnormal{(S1)}}{Attack}{Culinary arts developed over centuries may be lost.\hfill\hfill\textnormal{(S2)}}
In general, an argument from consequences may be decomposed into two parts: (i) whether $S$ is a positive consequence or a negative one; and (ii) whether the source of this consequence is consistent with or facilitated by $C$'s stance (S2), or is contrary to or obstructed by it (S1)\hide{\footnote{``Losing culinary arts'' is a consequence of ``stopping eating animal meat'', which is the claim's stance itself and hence ``consistent''. In contrast, ``a population with no empathy for other species'' is a consequence of ``the normalizing of killing animals for food'', which is contrary to the claim's stance.}}. 

Logically, $S$ is likely to support $C$ by presenting a positive (negative) consequence of a source that is consistent with (contrary to) $C$'s stance. In contrast, $S$ may attack $C$ by presenting a negative (positive) consequence of a source that is consistent with (contrary to) $C$'s stance. Given that $S$ describes consequence $Q$ of source $R$, this logic leads to:
\begin{resize}{10}
\begin{align*}
    &\textbf{R10:} &&\pred{BackingConseq}{S,C} \rightarrow \pred{Support}{S,C}, \\
    &\textbf{R11:} &&\pred{RefutingConseq}{S,C} \rightarrow \pred{Attack}{S,C}
\end{align*}\vspace{-6mm}
\begin{align*}
    \textit{s.t. } &\pred{BackingConseq}{S,C} = \\
    &~ P(S \text{ is a consequence}) \times \\
    &~~~ \left\{ P(Q \text{ is positive}) \cdot P(R \text{ consistent with } C) \right. \\
    &~~~~ + \left. P(Q \text{ is negative}) \cdot P(R \text{ contrary to } C) \right\}, \displaybreak[0] \\
    &\pred{RefutingConseq}{S,C} = \\
    &~ P(S \text{ is a consequence}) \times \\
    &~~~ \left\{ P(Q \text{ is negative}) \cdot P(R \text{ consistent with } C) \right. \\
    &~~~~ + \left. P(Q \text{ is positive}) \cdot P(R \text{ contrary to } C) \right\}.
\end{align*}
\end{resize}

\vspace{-7mm}
\paragraph{Practical Reasoning:}
In this scheme, the statement supports or attacks the claim by presenting a goal to achieve, as in:
\qqargu{Pregnant people should have the right to choose abortion.}{Support}{\ul{Women} should \ul{be able to make choices about their bodies}.\hfill\hfill\textnormal{(S3)}}{Attack}{Our rights do not allow us to \ul{harm the innocent lives of others}.\hfill\hfill\textnormal{(S4)}}
The statements use a normative statement as a goal to justify their stances. We call their target of advocacy or opposition (underlined above) a \textbf{norm target}. Generally, an argument of this scheme may be decomposed into: (i) whether $S$ advocates for its norm target (S3) or opposes it (S4), as if expressing positive or negative sentiment toward the norm target; and (ii) whether the norm target is a situation or action that is consistent with or facilitated by $C$'s stance, or that is contrary to or obstructed by it\footnote{Both harming innocent lives and making choices about their bodies are facilitated by the right to choose abortion (`consistent')}.

Logically, $S$ is likely to support $C$ by advocating for (opposing) a norm target that is consistent with (contrary to) $C$'s stance. In contrast, $S$ may attack $C$ by opposing (advocating for) a norm target that is consistent with (contrary to) $C$'s stance. Given that $S$ has norm target $R$, this logic leads to:
\begin{resize}{10}
\begin{align*}
    &\textbf{R12:} &&\pred{BackingNorm}{S,C} \rightarrow \pred{Support}{S,C}, \\
    &\textbf{R13:} &&\pred{RefutingNorm}{S,C} \rightarrow \pred{Attack}{S,C}
\end{align*}\vspace{-6mm}
\begin{align*}
    \textit{s.t. } &\pred{BackingNorm}{S,C} = \\
    &~ P(S \text{ is normative}) \times \\
    &~ \left\{ P(S \text{ advocates for } R) \cdot P(R \text{ consistent with } C) \right. \\
    &~~~ + \left. P(S \text{ opposes } R) \cdot P(R \text{ contrary to } C)  \right\}, \displaybreak[0] \\
    &\pred{RefutingNorm}{S,C} = \\
    &~ P(S \text{ is normative}) \times \\
    &~ \left\{ P(S \text{ opposes } R) \cdot P(R \text{ consistent with } C) \right. \\
    &~~~ + \left. P(S \text{ advocates for } R) \cdot P(R \text{ contrary to } C) \right\}.
\end{align*}
\end{resize}
The probabilities are computed by modules trained on our annotation data (\S{\ref{sec:annotation}}). 

\subsection{Relation Chain}
A chain of argumentative relations across arguments may provide information about the plausible relation within each argument. 
Given three statements $S$, $I$, and $C$, we have four chain rules:
\begin{resize}{9.5}
\begin{align*}
    &\textbf{R14: } \pred{Support}{S,I} \wedge \pred{Support}{I,C} \rightarrow \pred{Support}{S,C}, \displaybreak[0] \\
    &\textbf{R15: } \pred{Attack}{S,I} \wedge \pred{Attack}{I,C} \rightarrow \pred{Support}{S,C}, \displaybreak[0] \\
    &\textbf{R16: } \pred{Support}{S,I} \wedge \pred{Attack}{I,C} \rightarrow \pred{Attack}{S,C}, \displaybreak[0] \\
    &\textbf{R17: } \pred{Attack}{S,I} \wedge \pred{Support}{I,C} \rightarrow \pred{Attack}{S,C}. \displaybreak[0] 
\end{align*}
\end{resize}
For each data split, we combine two neighboring arguments where the claim of one is the statement of the other, whenever possible. The logical rules R1--R13 are applied to these ``indirect'' arguments.

\subsection{Constraints}
$C$ and $S$ are assumed to have the neutral relation (or the attack relation for binary classification) if they do not have strong evidence from the rules mentioned so far (Table \ref{tab:psl_rules} \textbf{C1}). In addition, the probabilities of all relations should sum to 1 (\textbf{C2}).

\section{Modules\label{sec:modules}}
In this section, we discuss individual modules for operationalizing the PSL rules.
\hide{Each module takes a text or a pair of texts as input and computes the probabilities of classes relevant to the module.} For each module, we fine-tune the pretrained uncased BERT-base~\cite{Devlin:2018bert}\hide{, which has shown great performance in many NLP tasks}. We use the transformers library v3.3.0~\cite{huggingface2020} for high reproducibility and low development costs. 
But any other models could be used instead.

Each dataset used is randomly split with a ratio of 9:1 for training and test. Cross-entropy and Adam are used for optimization. To address the imbalance of classes and datasets, the loss for each training instance is scaled by a weight inversely proportional to the number of its class and dataset.

\subsection{Textual Entailment\label{sec:mod_text_entail}}
A textual entailment module is used for rules about factual consistency and sentiment coherence (R1--R5). Given a pair of texts, it computes the probabilities of entailment, contradiction, and neutral. 

Our training data include two public datasets: MNLI~\cite{Williams:2018mnli} and AntSyn~\cite{Nguyen.2017antonyms} for handling antonyms and synonyms.
An NLI module combined with the word-level entailment handles short phrases better without hurting accuracy for sentence-level entailment.
Since AntSyn does not have the neutral class, we add 50K neutral word pairs by randomly pairing two words among the 20K most frequent words in MNLI; without them, a trained model can hardly predict the neutral relation between words. The accuracy for each dataset is in Table \ref{tab:modules} rows 1--4.

\setlength{\extrarowheight}{-1pt}
\begin{table}[t]
    \footnotesize
    \centering
    \begin{tabularx}{\linewidth}{@{~} p{5mm}p{0.8mm}p{4.5cm}r} \toprule
         & & Dataset (Classes, $N$) & Accuracy \\
        \midrule
        \rotaterow{4}{\fontsize{8}{6}\selectfont \multiline{Textual\\Entailment\\(R1--R5)}}
         & 1 & MNLI (ent/con/neu, 412,349) & F1=82.3 \\
         & 2 & AntSyn (ent/con, 15,632) & F1=90.2 \\
         & 3 & Neu50K (neu, 50,000) & R=97.5  \\
        \cmidrule(lr){2-4}
         & 4 & \textbf{MicroAvg (ent/con/neu, 477,981)} & \textbf{F1=84.7} \\
        \midrule
        \rotaterow{6}{\fontsize{8}{6}\selectfont \multiline{Sentiment\\Classification\\(R4--R5)}}
         & 5 & SemEval17 (pos/neg/neu, 20,632) & F1=64.5 \\
         & 6 & Dong (pos/neg/neu, 6,940) & F1=71.4 \\
         & 7 & Mitchell (pos/neg/neu, 3,288) & F1=62.5 \\
         & 8 & Bakliwal (pos/neg/neu, 2,624) & F1=69.7 \\
         & 9 & Norm (pos/neg, 632) & F1=100.0 \\
        \cmidrule(lr){2-4}
         & 10 & \textbf{MicroAvg (pos/neg/neu, 34,116)} & \textbf{F1=69.2} \\
        \midrule
        \rotaterow{9}{\fontsize{8}{6}\selectfont \multiline{Causality\\(R6--R9)}}
         & 11 & PDTB (cause/else, 14,224) & F1=68.1 \\ 
         & 12 & PDTB-R (cause/else 1,791) & F1=75.7 \\
         & 13 & BECauSE (cause/obstruct, 1,542) & F1=46.1 \\
         & 14 & BECauSE-R (else, 1,542) & R=86.5 \\
         & 15 & CoNet (cause, 50,420) & R=88.6 \\
         & 16 & CoNet-R (else, 50,420) & R=91.7 \\
         & 17 & WIQA (cause/obstruct, 31,630) & F1=88.2 \\
         & 18 & WIQA-P (else, 31,630) & R=90.2 \\
        \cmidrule(lr){2-4}
         & 19 & \shortstack{\textbf{MicroAvg (cause/obstr/else, 183,119)}} & \textbf{F1=87.7} \\ 
        \midrule
        \rotaterow{4}{\fontsize{8}{6}\selectfont \multiline{Normative\\Relation\\(R10--R13)}}
         & 20 & JustType (conseq/norm, 1,580) & F1=90.2 \\
         & 21 & ConseqSenti (pos/neg, 824) & F1=71.8 \\
         & 22 & NormType (adv/opp, 758) & F1=91.1 \\
         & 23 & $RC$-Rel (consist/contra/else, 1,924) & F1=70.1 \\
        \bottomrule
    \end{tabularx}
    \caption{F1-scores and recall of modules.}
    \label{tab:modules}
\end{table}
\setlength{\extrarowheight}{0pt}

\subsection{Target-Based Sentiment Classification\label{sec:mod_senti_analy}}
A sentiment classifier is for rules about sentiment coherence (R4--R5). Given a pair of texts $T_1$ and $T_2$, it computes the probability of whether $T_1$ has positive, negative, or neutral sentiment toward $T_2$.

Our training data include five datasets for target-based sentiment classification: SemEval17~\cite{semeval17task4},  entities~\cite{dong-etal-2014-adaptive}, open domain~\cite{mitchell-etal-2013-open}, Irish politics~\cite{Bakliwal:irish_senti}, and our annotations of positive/negative norms toward norm targets (\S{\ref{sec:annot_task1}}). These annotations highly improve classification of sentiments expressed through advocacy and opposition in normative statements. 
Pretraining on general sentiment resources---subjectivity lexicon~\cite{Wilson:2005subjlex} and sentiment140~\cite{Go:2009senti140}---also helps (Table \ref{tab:modules} rows 5--10).

\subsection{Causality\label{sec:mod_causality}}
\setlength{\extrarowheight}{-5pt}
\begin{table}[t]
    \fontsizeto{9}
    \centering
    \begin{tabularx}{\linewidth}{p{1.4cm}p{3.1cm}l} \toprule
        Corpus & Corpus-Specific Labels & Our Label ($N$) \\
        \midrule
        PDTB & Temporal.Asynchronous & Cause\hide{Precede} (1,255) \\
         & Temporal.Synchrnonous & Cause\hide{Sync} (536) \\
         & Comparison, Expansion & Else (12,433) \\
        PDTB-R$^\dagger$ & Temporal.Asynchronous & Else (536) \\
         & Temporal.Synchronous & Cause\hide{Sync} (1,255) \\
        \midrule
        BECauSE & Promote & Cause (1,417) \\
            & Inhibit & Obstruct (142) \\
        \shortstack{BECauSE-R$^\dagger$} & Promote, Inhibit & Else (1,613) \\
        \midrule
        WIQA & RESULTS\_IN & Cause (12,652) \\
            & NOT\_RESULTS\_IN & Obstruct (18,978) \\
        WIQA-P$^\ddagger$ & RESULSTS\_IN, NOT\_RESULTS\_IN & Else (31,630) \\
        \midrule
        ConceptNet & Causes, CausesDesire, & Cause (50,420) \hide{Cause (21,489)} \\
            & HasFirstSubevent, HasLastSubevent, HasPrerequisite & \hide{Precede (28,931)} \\
        ConceptNet-R$^\dagger$ & Causes, CausesDesire, HasFirstSubevent, HasLastSubevent, HasPrerequisite & Else (50,420) \\
        \bottomrule
    \end{tabularx}
    \caption{Mapping between corpus-specific labels and our labels for the causality module. $^\dagger$The order of two input texts are reversed. $^\ddagger$The second input text is replaced with a random text in the corpus.}
    \label{tab:causality_corpora}
\end{table}
\setlength{\extrarowheight}{0pt}
A causality module is used for rules regarding causal relations (R6--R9). Given an input pair of texts $T_1$ and $T_2$, it computes the probability of whether $T_1$ causes $T_2$, obstructs $T_2$, or neither.

Our training data include four datasets about causal and temporal relations between event texts. 
PDTB 3.0 \cite{Webber:2006pdtb3} is WSJ articles annotated with four high-level discourse relations, and we map the sub-relations of `Temporal' to our classes\footnote{We use explicit relations only for pretraining, since they often capture linguistically marked, rather than true, relations between events. We also exclude the Contingency relations as causal and non-causal relations (e.g., justification) are mixed.}.
BECauSE 2.0 \cite{Dunietz17:because} is news articles annotated with linguistically marked causality. 
WIQA \cite{Tandon19:wiqa} is scientific event texts annotated with causality between events. 
ConceptNet~\cite{Speer17:conceptnet} is a knowledge graph between phrases, and relations about causality are mapped to our classes.
To prevent overfitting to corpus-specific characteristics\hide{(e.g., genre, text length)}, we add adversarial data by swapping two input texts (PDTB-R, BECauSE-R, ConceptNet-R) or pairing random texts (WIQA-P).
The mapping between corpus-specific labels and ours is in Table \ref{tab:causality_corpora}, and the module accuracy in Table \ref{tab:modules} rows 11--19.

\subsection{Normative Relation\label{sec:mod_schemes}}
All the modules here are trained on our annotations of normative argumentation schemes (\S{\ref{sec:annotation}}).

\paragraph{$\boldsymbol{P(S \text{ is a consequence / norm})}$ (R10--R13):}
Given a statement, one module computes the probability that it is a consequence, and another module the probability of a norm. Both modules are trained on all claims and statements in our annotations, where all claims are naturally norms, and each statement is annotated as either norm or consequence (Table~{\ref{tab:modules}} row 20).

\paragraph{$\boldsymbol{P(Q \text{ is positive / negative})}$ (R10--R11):}
Given a statement assumed to be a consequence, this module computes the probability of whether it is positive or negative. It is trained on all statements annotated as consequence (Table~\ref{tab:modules} row 21). 

\paragraph{$\boldsymbol{P(S \text{ advocates / opposes})}$ (R12--R13):}
Given a statement assumed to be a norm, this module computes the probability of whether it is advocacy or opposition. It is trained on all claims, plus statements annotated as norm (Table~\ref{tab:modules} row 22).

\paragraph{$\boldsymbol{P(R \text{ consistent / contrary to } C)}$ (R10--R13):}
For a pair of $S$ and $C$, the module computes the probability of whether $R$ (the norm target or the source of consequence in $S$) and $C$'s stance are consistent, contrary, or else. In our annotations, $R$ and $C$ are `consistent' if both (1a and 3a in Figure \ref{fig:annotation_overall}) are advocacy or opposition, and `contrary' otherwise. To avoid overpredicting the two classes, we add negative data by pairing $C$ with a random statement in the annotations. The module is pretrained on MNLI and AntSyn (Table~\ref{tab:modules} row 23).

\section{Annotation of Normative Argumentation Schemes\label{sec:annotation}}
\begin{figure}[t]
    \centering
    \includegraphics[width=.95\linewidth]{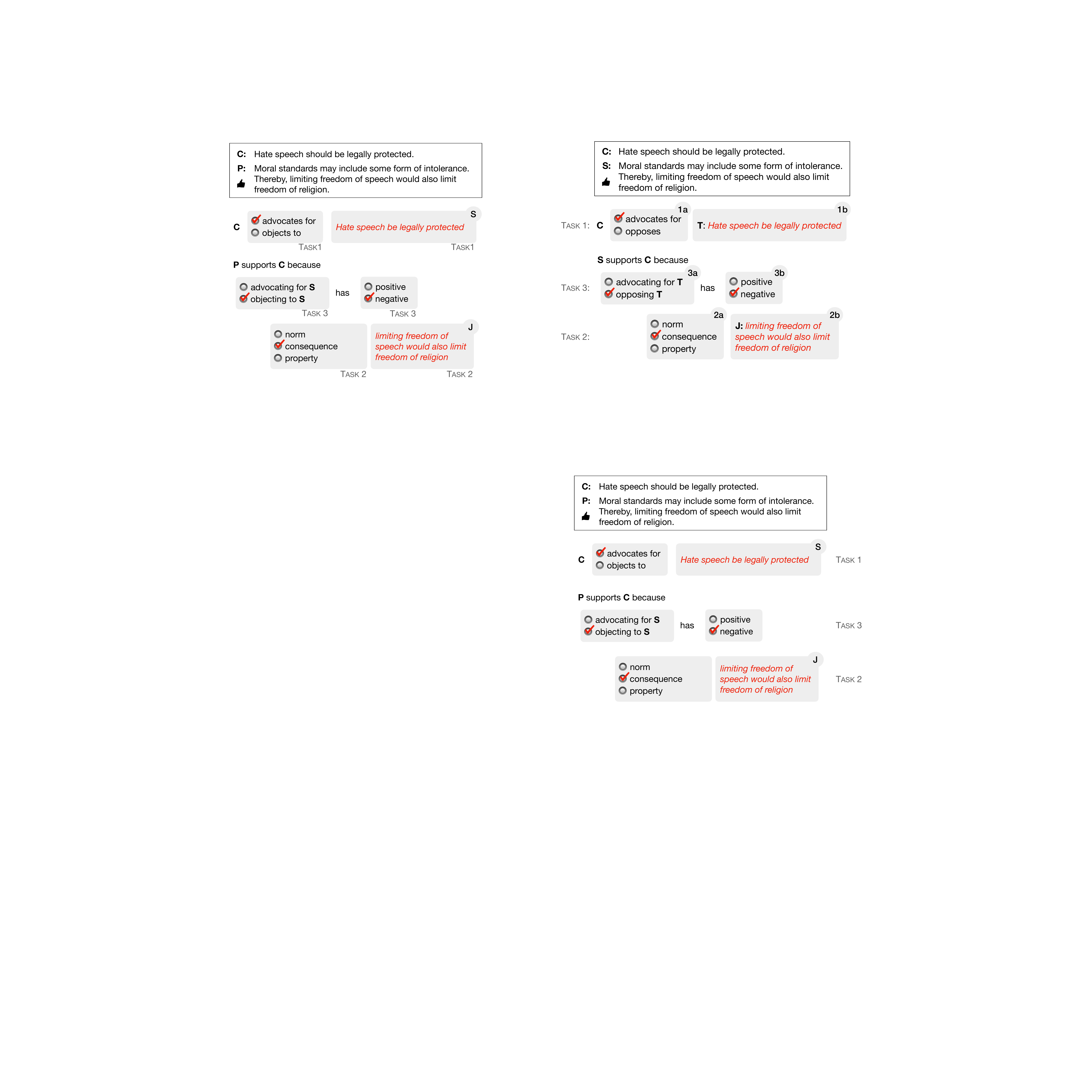}
    \caption{Example annotations (checks and italic) of the normative argumentation schemes. It depends on the argument whether $S$ supports or attacks $C$.}
    \label{fig:annotation_overall}
\end{figure}

In this section, we discuss our annotation of the argumentation schemes \textit{argument from consequences} and \textit{practical reasoning} (Figure \ref{fig:annotation_overall}). The resulting annotations are used to train the modules in \S{\ref{sec:mod_schemes}} which compute the probability terms in R10--R13.

For each pair of normative claim $C$ and statement $S$, we annotate the following information:
(1a) Whether $C$ advocates for or opposes its norm target, and (1b) the norm target $T$ (Figure \ref{fig:annotation_overall} TASK 1); (2a) Whether $S$ uses a norm, consequence, or property for justification, and (2b) the justification $J$ (Figure \ref{fig:annotation_overall} TASK 2); (3a) Whether $J$'s focus is on advocating for $T$ or opposing $T$, and (3b) whether $J$ is positive or negative (Figure \ref{fig:annotation_overall} TASK 3).\footnote{This annotation schema provides enough information for the classifiers in \S{\ref{sec:mod_schemes}}. $P$($S$ is a consequence / norm) is from (2a), and both $P$($Q$ is positive / negative) and $P$($S$ advocates / opposes) are from (3b). $P$($R$ consistent / contrary to $C$) can be obtained by combining (1a) and (3a): `consistent' if both advocate or both oppose, and `contrary' otherwise.}

Our annotation schema is richer than existing ones \cite{Lawrence:2016kt,Reisert.2018.schemes}. Due to the increased complexity, however, our annotation is split into three pipelined tasks. For this annotation, we randomly sampled 1,000 arguments from Kialo whose claims are normative (see \S{\ref{sec:data}} and Table \ref{tab:datasets} for details).

\subsection{Task 1. Norm Type/Target of Claim\label{sec:annot_task1}}
For each $C$, we annotate: (1a) the norm type---advocate, oppose, or neither---toward its norm target; and (1b) the norm target $T$. 
Advocacy is often expressed as ``should/need T'', whereas opposition as ``should not T'', ``T should be banned''; `neither' is noise (2.8\%) to be discarded. $T$ is annotated by rearranging words in $C$ (Figure \ref{fig:annotation_overall} TASK 1).

There are 671 unique claims in the annotation set. The first author of this paper wrote an initial manual and trained two undergraduate students majoring in economics, while resolving disagreements through discussion and revising the manual. In order to verify that the annotation can be conducted systematically, we measured inter-annotator agreement (IAA) on 200 held-out claims. The annotation of norm types achieved Krippendorff's $\alpha$ of 0.81\hide{ (95\% CI=(0.74, 0.88) with the bootstrap)}. To measure IAA for annotation of $T$, we first aligned words between each annotation and the claim\hide{\footnote{We iteratively matched and excluded the longest common substring.}}, obtaining a binary label for each word in the claim (1 if included in the annotation). As a result, we obtained two sequences of binary labels of the same length from the two annotators and compared them, achieving an F1-score of 0.89\hide{ (95\% CI=(0.86, 0.91))}. The high $\alpha$ and F1-score show the validity of the annotations and annotation manual. All disagreements were resolved through discussion afterward.\footnote{These annotations are used for the sentiment classifiers in \S{\ref{sec:mod_senti_analy}} too. For example, ``the lottery should be banned'' is taken to express negative sentiment toward the lottery. Such examples are underrepresented in sentiment datasets, resulting in inaccurate sentiment classification for normative statements.}

\subsection{Task 2. Justification Type of Premise\label{sec:annot_task2}}
For each pair of $C$ and $S$, we annotate: (2a) the justification type of $S$---norm, consequence, property, or else; and (2b) the justification $J$. The justification types are defined as follows:
\begin{itemize}
    \item \textbf{Norm:} $J$ states that some situation or action should be achieved (practical reasoning).
    \item \textbf{Consequence:} $J$ states a potential or past outcome (argument from consequences).
    \item \textbf{Property:} $J$ states a property that (dis)quali-fies $C$'s stance (argument from consequence).
\end{itemize}
The difference between consequence and property is whether the focus is on extrinsic outcomes or intrinsic properties, such as feasibility, moral values, and character (e.g., ``Alex shouldn't be the team leader because he is dishonest''). 
We consider both as argument from consequences because property-based justification has almost the same logic as consequence-based justification. The `else' type is rare (3.4\%) and discarded after the annotation.

The process of annotation and IAA measurement is the same as Task 1, except that IAA was measured on 100 held-out arguments due to a need for more training. 
For justification types, Krippendorff's $\alpha$ is 0.53\hide{ with 95\% CI=(0.41, 0.65)}---moderate agreement. For justification $J$, the F1-score is 0.85\hide{ with 95\% CI=(0.80, 0.90)}. 
The relatively low IAA for justification types comes from two main sources. First, a distinction between consequence and property is fuzzy by nature, as in ``an asset tax is the most fair system of taxing citizens''. 
This difficulty has little impact on our system, however, as both are treated as argument from consequences.
\hide{If we combine these two categories, Krippendorff's $\alpha$ increases to 0.58 with 95\% CI=(0.37, 0.77).}
Second, some statements contain multiple justifications of different types. If so, we asked the annotators to choose one that they judge to be most important (for training purposes). They sometimes chose different justifications, although they usually annotated the type correctly for the chosen one.
\hide{Lastly, since the `else' type is rare, disagreements on it hurt IAA significantly.}

\subsection{Task 3. Justification Logic of Statement\label{sec:annot_task3}}
Given $C$ with its norm target $T$, and $S$ with its justification $J$, we annotate: (3a) whether the consequence, property, or norm target of $J$ is regarding advocating for $T$ or opposing $T$; and (3b) whether $J$ is positive or negative. $J$ is positive (negative) if it's a positive (negative) consequence/property or expresses advocacy (opposition).

\hide{In the following argument
\qargu{People should \ul{eat whatever they feel like eating}.}{}{There is no reason to \ul{deny oneself a pleasure}. \textnormal{(norm targets underlined)}}
$S$ is a \textit{negative} norm because it is opposing its norm target, by saying ``there is no reason to''. This norm target is consistent with \textit{opposing} the claim's norm target $T$, because denying oneself a pleasure is contrary to eating whatever they feel like eating.}

\hide{Some ambiguous cases include:
\qargu{The roles an actor can play should be limited by that actor’s race, gender, or sexuality.}{Attack}{An actor's talent may be more important than getting an exact visual match in some cases.}
Here, the statement may be considered positive or negative depending on the perspective of talent or race. In this case, we annotate the overall sentiment of the statement, which in this case is positive reflected in ``more important''.}

This task was easy, so only one annotator worked with the first author. Their agreement measured on 400 heldout arguments is Krippendorff's $\alpha$ of 0.82\hide{(95\% CI=(0.77, 0.86))} for positive/negative and 0.78\hide{(95\% CI=(0.72, 0.83))} for advocate/oppose.

\subsection{Analysis of Annotations\label{sec:annot_analy}}
We obtained 962 annotated arguments with claims of advocacy (70\%) and opposition (30\%), and statements of consequence (54\%), property (32\%), and norm (14\%).
Supporting statements are more likely to use a positive justification (62\%), while attacking statements a negative one (68\%), with significant correlations ($\chi^2=87, p<.00001$). 
But 32--38\% of the time, they use the opposite sentiment, indicating that sentiment alone cannot determine argumentative relations.
\hide{Supporting statements tend to emphasize the positivity of what the claim advocates for (74\%) or the negativity of what the claim opposes (66\%). While attacking statements often emphasize the negativity of what the claim advocates for (76\%), positivity and negativity are equally emphasized (50\%) for claims that show opposition.}
\hide{Statements tend to present a direct indication (consequence or norm) of the claim's stance rather than an indication of the opposite of the claim's stance, while attacking statements are more likely so (68\%) than supporting statements (60\%) ($\chi^2=5.9, p<.05$). Especially when attacking claims that advocate for something, statements tend bring up direct negativity of it (76\%).}

\section{Data\label{sec:data}}

\paragraph{Kialo:} Our first dataset is from kialo.com, a collaborative argumentation platform covering contentious topics. Users contribute to the discussion of a topic by creating a statement that either supports or attacks an existing statement, resulting in an argumentation tree for each topic. We define an \textbf{argument} as a pair of parent and child statements, where the parent is the \textbf{claim} and the child is the \textbf{support or attack statement}. Each argument is labeled with support or attack by users and is usually self-contained, not relying on external context, anaphora resolution, or discourse markers.

We scraped arguments for 1,417 topics\hide{written until Oct 2019,} and split into two subsets. \textbf{Normative arguments} have normative claims suggesting that a situation or action be brought about, while \textbf{non-normative arguments} have non-normative claims. 
This distinction helps us understand the two types of arguments better. We separated normative and non-normative claims using a BERT classifier trained on \citet{Jo:2020lrec}'s dataset of different types of statements (AUC=98.8\%), as binary classification of normative statement or not. A claim is considered normative (non-normative) if the predicted probability is higher than 0.97 (lower than 0.4); 
claims with probability scores between these thresholds (total 10\%) are discarded to reduce noise.

In practice, an argument mining system may also need to identify statements that seem related but do not form any argument.
Hence, we add the same number of ``neutral arguments'' by pairing random statements within the same topic.
To avoid paired statements forming a reasonable argument accidentally,
we constrain that they be at least 9 statements apart in the argumentation tree, making them unlikely to have any support or attack relation but still topically related to each other.

Among the resulting arguments, 10K are reserved for fitting; 20\% or 30\% of the rest (depending on the data size) are used for validation and the others for test (Table~\ref{tab:datasets}). We increase the validity of the test set by manually discarding non-neutral arguments from the neutral set. We also manually inspect the normativity of claims, and if they occur in the fitting or validation sets too, the corresponding arguments are assigned to the correct sets according to the manual judgments. For normative arguments, we set aside 1,000 arguments for annotating the argumentation schemes (\S{\ref{sec:annotation}}). 

The data cover the domains economy (13\%), family (11\%), gender (10\%), crime (10\%), rights (10\%), God (10\%), culture (10\%), entertainment (7\%), and law (7\%), as computed by LDA. The average number of words per argument is 49 (45) for normative (non-normative) arguments.

\newcolumntype{P}[1]{>{\raggedleft\arraybackslash}p{#1}}
\newcolumntype{R}{>{\raggedleft\arraybackslash}X}

\begin{table}[t]
    \centering
    \footnotesize
    \begin{tabularx}{\linewidth}{p{2mm}P{2.5mm}P{4mm}P{6.5mm}P{6.5mm}P{6.5mm}P{5mm}P{3mm}P{3mm}} 
        \toprule
         & & \multicolumn{4}{c}{Kialo} & \multicolumn{3}{c}{Debatepedia} \\
         \cmidrule{3-6} \cmidrule(l){7-9}
         & & {\fontsize{8}{0}\selectfont\multiline{Annot-\\ation}} & \shortstack{Fit} & \shortstack{Val} & \shortstack{Test} & \shortstack{Fit} & \shortstack{Val} & \shortstack{Test}  \\
        \midrule
        \rotaterow{3}{\fontsize{8}{2}\selectfont\multiline{\\Normative}} & Sup & 480 & 4,621 & 1,893 & ~~6,623 & 6,598 & 229 & 356\\
         & Att & 520 & 5,383 & 2,124 & ~~7,623  & 4,502 & 243 & 351\\
         & Neu & -- & 9,984 & 4,000 & 14,228 & -- & -- & -- \\
        \midrule
        \rotaterow{3}{\fontsize{8.5}{0}\selectfont\multiline{Non-\\normative}} & Sup & -- & ~~4,953 & 10,135 & 21,138 & 3,302 & 243 & 178 \\
         & Att & -- & ~~5,043 & ~~9,848 & 20,197 & 3,278 & 253 & 152 \\
         & Neu & -- & 10,016 & 20,000 & 40,947 & -- & -- & -- \\
        \bottomrule
    \end{tabularx}
    \caption{Numbers of arguments in datasets.}
    \label{tab:datasets}
\end{table}

\paragraph{Debatepedia:} The second dataset is Debatepedia arguments \cite{Hou.2017}. 508 topics are paired with 15K pro and con responses, and we treat each pair as an \textbf{argument} and each topic and response as \textbf{claim} and \textbf{statement}, respectively. 

One important issue is that most topics are in question form, either asking if you agree with a stance (``yes'' is pro and ``no'' is con) or asking to choose between two options (the first is pro and the second is con). 
Since our logical mechanisms do not handle such questions naturally, we convert them to declarative claims as follows. 
The first type of questions are converted to a claim that proposes the stance (e.g., ``Should Marijuana be legalized?'' to ``Marijuana should be legalized''), and the second type of questions to a claim that prefers the first option (e.g., ``Mission to the Moon or Mars?'' to ``Mission to the Moon is preferred to Mars''). The first author and an annotator converted all topics independently and then resolved differences. 

We split the arguments into \textbf{normative} and \textbf{non-normative} sets as we do for Kialo, manually verifying all claims. There is no neutral relation. We use the original train, validation, and test splits (Table~\ref{tab:datasets}). Debatepedia claims are shorter and less diverse than Kialo claims. They focus mostly on valuation, while Kialo includes a lot of factual claims.

\section{Experiment 1. Probabilistic Soft Logic\label{sec:exp1}}

The goal here is to see how well the logical mechanisms alone can explain argumentative relations. 

\subsection{PSL Settings\label{sec:exp1_settings}}
We use the PSL toolkit v2.2.1\footnote{\url{https://psl.linqs.org/wiki/2.2.1/}}. The initial weights of the logical rules R1--R13 are set to 1. 
The importance of the chain rules R14--R17 may be different, so we explore \{1, 0.5, 0.1\}. 
The weight of C1 serves as a threshold for the default relation (i.e., neutral for Kialo and attack for Debatepedia), and we explore \{0.2, 0.3\}; initial weights beyond this range either ignore or overpredict the default relation. C2 is a hard constraint. The optimal weights are selected by the objective value on the validation set (this does not use true relation labels).

\subsection{Baselines\label{sec:exp1_baselines}}
We consider three baselines. \textbf{Random} assigns a relation to each argument randomly. 
\textbf{Sentiment} assigns a relation based on the claim and statement's agreement on sentiment: support if both are positive or negative, attack if they have opposite sentiments, and neutral otherwise. 
\hide{This generally outperforms labeling based on the statement's sentiment only.}
We compute a sentiment distribution by averaging all target-specific sentiments from our sentiment classifier (\S{\ref{sec:mod_senti_analy}}). \textbf{Textual entailment} assigns support (attack) if the statement entails (contradicts) the claim, and neutral otherwise \cite{Cabrio:2012ud}. We use our textual entailment module (\S{\ref{sec:mod_text_entail}}). For Debatepedia, we choose between support and attack whichever has a higher probability.

\subsection{Results\label{sec:exp1_results}}
\setlength{\extrarowheight}{-10pt}
\begin{table*}[t]
    \centering
    \fontsizeto{8}
    \begin{subtable}[t]{.85\linewidth}
        \begin{tabularx}{\linewidth}{@{~} p{1mm}l XXXXXX p{2mm} XXXXXX} \toprule
             & & \multicolumn{6}{c}{Normative Arguments}  & 
                & \multicolumn{6}{c}{Non-normative Arguments} \\
            \cmidrule{3-8} \cmidrule(l){10-15}
             & & ACC & AUC & F1 & \shortstack{\Fsup} & \shortstack{\Fatt} & \shortstack{\Fneu}   &
                & ACC & AUC & F1 & \shortstack{\Fsup} & \shortstack{\Fatt} & \shortstack{\Fneu} \\
            \midrule
            1 & Random & 33.5 & 50.2 & 32.6 & 27.8 & 30.1 & 39.9 &
                & 33.4 & 49.9 & 32.5 & 28.7 & 28.8 & 40.0 \\
            2 & Sentiment & 40.8 & 64.1 & 40.7 & 40.6 & 39.1 & 42.4 &
                 & 43.7 & 61.1 & 42.2 & 40.0 & 35.2 & 51.5 \\
            3 & Text Entail & 51.8 & 61.8 & 36.7 & 12.8 & 30.4 & 67.0 &
                 & 52.1 & 62.8 & 38.6 & 18.4 & 31.0 & 66.4 \\
            \midrule
            4 & PSL (R1--R13) & 54.0$^\ddagger$ & 73.8$^\ddagger$ & 52.1$^\ddagger$ & 47.0$^\ddagger$ & 43.6$^\ddagger$ & 65.7$^\ddagger$  &
                & 57.0$^\ddagger$ & 76.0$^\ddagger$ & 54.0$^\ddagger$ & 50.1$^\ddagger$ & 42.6$^\ddagger$ & 69.3$^\ddagger$ \\
            5 & $\backslash$ Fact & 55.1$^\ddagger$ & 74.3$^\ddagger$ & 52.4$^\ddagger$ & 47.1$^\ddagger$ & 41.6$^\ddagger$ & 68.4$^\ddagger$ &
                & 58.6$^\ddagger$ & 77.1$^\ddagger$ & 55.1$^\ddagger$ & 50.5$^\ddagger$ & 42.2$^\ddagger$ & 72.7$^\ddagger$ \\
            6 & $\backslash$ Sentiment & \textbf{62.1}$^\ddagger$ & 77.6$^\ddagger$ & 57.5$^\ddagger$ & 49.1$^\ddagger$ & 45.8$^\ddagger$ & \textbf{77.7}$^\ddagger$ &
                & 61.3$^\ddagger$ & 77.8$^\ddagger$ & 56.7$^\ddagger$ & 50.3$^\ddagger$ & 44.1$^\ddagger$ & 75.7$^\ddagger$ \\
            7 & $\backslash$ Causal & 54.4$^\ddagger$ & 73.1$^\ddagger$ & 52.3$^\ddagger$ & 45.4$^\ddagger$ & 45.4$^\ddagger$ & 66.0$^\ddagger$  &
                & 57.6$^\ddagger$ & 76.1$^\ddagger$ & 54.3$^\ddagger$ & 48.7$^\ddagger$ & 43.4$^\ddagger$ & 70.7$^\ddagger$ \\
            8 & $\backslash$ Normative & 51.8$^\ddagger$ & 68.6$^\ddagger$ & 49.4$^\ddagger$ & 44.3$^\ddagger$ & 40.4$^\dagger$ & 63.4$^\ddagger$ & 
                & 54.7$^\ddagger$ & 70.3$^\ddagger$ & 51.4$^\ddagger$ & 47.0$^\ddagger$ & 40.3$^\ddagger$ & 66.8$^\ddagger$ \\
            \midrule
            9 & $\backslash$ Sentiment + Chain & 61.9$^\ddagger$ & \textbf{77.7}$^\ddagger$ & \textbf{57.7}$^\ddagger$ & \textbf{49.3}$^\ddagger$ & \textbf{46.2}$^\ddagger$ & 77.6$^\ddagger$  &
                & \textbf{61.5}$^\ddagger$ & \textbf{78.0}$^\ddagger$ & \textbf{57.2}$^\ddagger$ & \textbf{50.8}$^\ddagger$ & \textbf{44.7}$^\ddagger$ & \textbf{76.1}$^\ddagger$  \\
            \bottomrule
        \end{tabularx}
        \caption{Kialo}
        \label{tab:accs_psl_kialo}
    \end{subtable}
    \begin{subtable}[t]{.7\linewidth}
        \begin{tabularx}{\linewidth}{@{~} p{1mm}l XXXXX p{2mm} XXXXX} \toprule
             & & \multicolumn{5}{c}{Normative Arguments} & 
                & \multicolumn{5}{c}{Non-normative Arguments} \\
            \cmidrule{3-7} \cmidrule(l){9-13}
             & & ACC & AUC & F1 & \shortstack{\Fsup} & \shortstack{\Fatt} & 
                & ACC & AUC & F1 & \shortstack{\Fsup} & \shortstack{\Fatt} \\
            \midrule
            1 & Random & 47.7 & 49.4 & 50.2 & 49.0 & 51.4 &
                & 53.0 & 54.6 & 52.4 & 53.7 & 51.1 \\
            2 & Sentiment & 59.3 & 63.9 & 59.2 & 61.0 & 57.4 &
                & 69.1 & 73.4 & 68.5 & 72.7 & 64.3 \\
            3 & Text Entail & 52.2 & 55.8 & 49.4 & 37.6 & 61.2 &
                & 70.6 & 74.2 & 70.5 & 69.0 & 72.0 \\
            \midrule
            4 & PSL (R1--R13) & \textbf{63.9}$^\star$ & \textbf{68.3}$^\star$ & \textbf{63.9}$^\star$ & 63.8 & 64.0$^\dagger$ &
                & 73.0 & 76.1 & 73.0 & 74.2 & 71.7 \\
            5 & $\backslash$ Fact & 63.4$^\star$ & 67.1 & 63.4$^\star$ & \textbf{64.0} & 62.7$^\star$ &
                & 71.8 & 75.6 & 71.7 & 73.2 & 70.3 \\
            6 & $\backslash$ Sentiment & 63.1$^\star$ & 67.2 & 63.1$^\star$ & 62.7 & 63.5$^\star$ &
                & 70.9 & 74.0 & 70.9 & 71.6 & 70.2 \\
            7 & $\backslash$ Causal & 62.4$^\star$ & 66.3 & 62.1$^\star$ & 58.6 & \textbf{65.5}$^\star$ &
                & \textbf{74.5} & \textbf{78.7} & \textbf{74.5} & \textbf{75.4} & \textbf{73.6} \\
            8 & $\backslash$ Normative & 61.0 & 64.7 & 61.0 & 60.3 & 61.6$^\star$ &
                & 68.2 & 72.4 & 68.2 & 68.3 & 68.1 \\
            \bottomrule
        \end{tabularx}
        \caption{Debatepedia}
        \label{tab:accs_psl_debate}
    \end{subtable}
    \caption{PSL accuracy. $p$ < \{0.05$^\star$, 0.01$^\dagger$, 0.001$^\ddagger$\} with paired bootstrap compared to the best baseline.}
        \label{tab:accs_psl}
\end{table*}
\setlength{\extrarowheight}{0pt}

Tables \ref{tab:accs_psl_kialo} and \ref{tab:accs_psl_debate} summarize the accuracy of all models for Kialo and Debatepedia, respectively.  Among the baselines, sentiment (row 2) generally outperforms textual entailment (row 3), both significantly better than random (row 1). Sentiment tends to predict the support and attack relations aggressively, missing many neutral arguments, whereas
textual entailment is conservative and misses many support and attack arguments.
PSL with all logical rules R1--R13 (row 4) significantly outperforms all the baselines with high margins, and its F1-scores are more balanced across the relations. 

To examine the contribution of each logical mechanism, we conducted ablation tests (rows 5--8). The most contributing mechanism is clearly normative relation across all settings, without which F1-scores drop by 2.6--4.8 points (row 8).
This indicates that our operationalization of \textit{argument from consequences} and \textit{practical reasoning} can effectively explain a prevailing mechanism of argumentative relations. 

Quite surprisingly, normative relation is highly informative for non-normative arguments as well for both datasets. 
To understand how this mechanism works for non-normative arguments, we analyzed arguments for which it predicted the correct relations with high probabilities. 
It turns out that even for non-normative claims, the module often interprets negative sentiment toward a target as an opposition to the target. For the following example,
\qargu{Schooling halts individual development.}{Attack}{Schooling, if done right, can lead to the development of personal rigor ...}
the module implicitly judges the ``schooling'' in the claim to be opposed and thus judges the ``schooling'' in the statement (the source of consequence) to be contrary to the claim's stance while having positive sentiment (i.e., R11 applies).
This behavior is reasonable, considering how advocacy and opposition are naturally mapped to positive and negative norms in our annotation schema (\S{\ref{sec:annot_task3}}).

The utility of normative relation for non-normative arguments is pronounced for Debatepedia. Excluding this mechanism leads to a significant drop of F1-scores by 4.8 points (Table~\ref{tab:accs_psl_debate} row 8). One possible reason is that most claims in the non-normative set of Debatepedia are valuation; that is, they focus on whether something is good or bad, or preferences between options. As discussed above, valuation can be handled by this mechanism naturally. And in such arguments, causal relation may provide only little and noisy signal (row 7).

Sentiment coherence is the second most contributing mechanism. 
For Kialo, including it in the presence of normative relation is rather disruptive (Table~\ref{tab:accs_psl_kialo} row 6). This may be because the two mechanisms capture similar (rather than complementary) information, but sentiment coherence provides inaccurate information conflicting with that captured by normative relation.
Without normative relation, however, sentiment coherence contributes substantially more than factual consistency and causal relation by 4.4--5.9 F1-score points (not in the table). For Debatepedia, the contribution of sentiment coherence is clear even in the presence of normative relation (Table~\ref{tab:accs_psl_debate} row 6).

Factual consistency and causal relation have high precision and low recall for the support and attack relations. This explains why their contribution is rather small overall and even obscure for Kialo in the presence of normative relation (Table~\ref{tab:accs_psl_kialo} rows 5 \& 7). However, without normative relation they contribute 0.7--1.1 F1-score points for Kialo (not in the table). For Debatepedia, factual consistency contributes 0.5--1.3 points (Table~\ref{tab:accs_psl_debate} row 5), and causal relation 1.8 points to normative arguments (row 7).
Their contributions show different patterns in a supervised setting, however, as discussed in the next section.

To apply the chain rules (R14--R17) for Kialo, we built 16,328 and 58,851 indirect arguments for the normative and non-normative sets, respectively. Applying them further improves the best performing PSL model (Table~\ref{tab:accs_psl_kialo} row 12). It suggests that there is a relational structure among arguments, and structured prediction can reduce noise in independent predictions for individual arguments.

There is a notable difference in the performance of models between the three-class setting (Kialo) and the binary setting (Debate). The binary setting makes the problem easier for the baselines, reducing the performance gap with the logical mechanisms. When three relations are considered, the sentiment baseline and the textual entailment baseline suffer from low recall for the neutral and support/attack relations, respectively. But if an argument is guaranteed to belong to either support or attack, these weaknesses seem to disappear.

\subsection{Error Analysis\label{sec:exp1_error_analy}}

We conduct an error analysis on Kialo.
For the mechanism of normative relation, we examine misclassifications in normative arguments by focusing on the 50 support arguments and 50 attack arguments with the highest probabilities of the opposite relation.
Errors are grouped into four types: $R$-$C$ consistency/contrary (60\%), consequence sentiment (16\%), ground-truth relation (8\%), and else (16\%). 
The first type is mainly due to the model failing to capture antonymy relations, such as \lex{collective presidency $\leftrightarrow$ unitary presidency} and \lex{marketplace of ideas $\leftrightarrow$ deliver the best ideas}. Integrating advanced knowledge may rectify this issue. The second type of error often arises when a statement has both positive and negative words, as in ``student unions could \textit{prevent} professors from \textit{intentionally failing students} due to personal factors''.

For the other mechanisms, we examine non-normative arguments that each mechanism judged to have strong signal for a false relation. To that end, for each predicate in R1--R9, we choose the top 20 arguments that have the highest probabilities but were misclassified. 
Many errors were simply due to the misclassification of the classification modules, which may be rectified by improving the modules' accuracy. 
But we also found some blind spots of each predicate. For instance, FactEntail often fails to handle concession and scoping. 
\qargu{Fourth wave feminists espouse belief in equality.}{Attack}{It is belief in equality of outcome not opportunity that fourth wave feminists are espousing with quotas and beneficial bias.}
For SentiConsist, a statement can have the same ground of value as the claim without supporting it:
\qargu{The education of women is an important objective to \ul{improve the overall quality of living}.}{Attack}{Education of both men and women will have greater effects than that of women alone. Both must play a role in \ul{improving the quality of life of all of society's members}.}
The statement attacks the claim while expressing the same sentiment toward the same target (underlined).

\section{\mbox{Experiment 2. Representation Learning}\label{sec:exp2}}
\hide{We have seen that factual consistency, sentiment coherence, causal relation, and normative relation are working mechanisms in argumentation and can predict argumentative relations to a certain degree without training on relation-labeled data. However, argumentative relations have correlations with other statistics as well, such as thematic associations between a topic and a stance (framing), use of negating words, and sentiment, and supervised models are good at leveraging them, even excessively sometimes~\cite{Niven:2019cq,Allaway:2020stance}.}
Supervised models are good at capturing various associations between argumentative relations and data statistics.
Here, we examine if our logical me-chanisms can further inform them. We describe a simple but effective representation learning method, followed by baselines and experiment results.

\subsection{Method}
\hide{We present a simple method here that aims to learn a representation of the input argument that is ``cognizant of'' our logical mechanisms.}
Our logical mechanisms are based on textual entailment, sentiment classification, causality classification, and four classification tasks for normative relation (\S{\ref{sec:modules}}). We call them \textbf{logic tasks}. 
We combine all minibatches across the logic tasks using the same datasets from \S{\ref{sec:modules}} except the heuristically-made negative datasets. Given uncased BERT-base, we add a single classification layer for each logic task and train the model on the minibatches for five epochs in random order.
After that, we fine-tune it on our fitting data (Table \ref{tab:datasets}), where the input is the concatenation of statement and claim. Training stops if AUC does not increase for 5 epochs on the validation data. We call our model \textbf{LogBERT}\hide{, while the base model does not have to be BERT}. 
\hide{To avoid catastrophic forgetting, we tried blending the logic and main tasks~\cite{Shnarch:2018blend} and using regularized fine-tuning~\cite{Aghajanyan:2020r3f}, but they did not help.}

\begin{table*}[t]
    \centering
    \fontsizeto{8}
    \begin{subtable}[t]{.75\linewidth}
        \begin{tabularx}{\linewidth}{@{~} p{1mm}lXXXXXX p{3mm} XXXXXX} \toprule
             &  & \multicolumn{6}{c}{Normative Arguments} 
                & & \multicolumn{6}{c}{Non-normative Arguments} \\
            \cmidrule{3-8} \cmidrule{10-15}
             &  & ACC & AUC & F1 & \shortstack{\Fsup} & \shortstack{\Fatt} & \shortstack{\Fneu} 
                & & ACC & AUC & F1 & \shortstack{\Fsup} & \shortstack{\Fatt} & \shortstack{\Fneu} \\
            \midrule
            1 & TGA Net & 71.5 & 88.3 & 62.2 & 43.5 & 54.3 & 88.7 
                & & 76.6 & 90.8 & 69.8 & 62.9 & 53.9 & 92.5 \\
            2 & Hybrid Net & 66.8 & 78.2 & 56.2 & 42.9 & 42.4 & 83.4 
                & & 71.8 & 82.2 & 65.7 & 55.6 & 51.4 & 90.2 \\
            \midrule
            3 & BERT & 79.5 & 92.4 & 73.3 & 60.5 & 65.2 & \textbf{94.2} 
                & & 83.8 & 94.6 & 79.2 & 72.3 & 68.8 & 96.6 \\
            4 & BERT+LX & 79.2 & 92.1 & 72.7 & 58.7 & 65.6$^\star$ & 93.8 
                & & 83.7 & 94.6 & 79.2 & 70.8 & 69.9$^\ddagger$ & \textbf{96.9}$^\ddagger$ \\
            5 & BERT+MT & 79.3 & 92.6$^\star$ & 73.4 & \textbf{63.8}$^\ddagger$ & 63.6 & 92.7 
                & & 83.6 & 94.7 & 79.2 & 71.8 & 69.7$^\ddagger$ & 96.1 \\
            \midrule
            6 & LogBERT & \textbf{80.0}$^\ddagger$ & \textbf{92.8}$^\ddagger$ & \textbf{74.3}$^\ddagger$ & 63.6$^\ddagger$ & \textbf{66.2}$^\ddagger$ & 93.2 
                & & \textbf{84.3}$^\ddagger$ & \textbf{95.0}$^\ddagger$ & \textbf{80.2}$^\ddagger$ & \textbf{73.1}$^\ddagger$ & \textbf{71.4}$^\ddagger$ & 96.1 \\
            \bottomrule
        \end{tabularx}
        \caption{Kialo}
        \label{tab:accs_sup_kialo}
    \end{subtable}
    \begin{subtable}[t]{.65\linewidth}
        \begin{tabularx}{\linewidth}{@{~} p{1mm}lXXXXX p{2mm} XXXXX} \toprule
             &  & \multicolumn{5}{c}{Normative Arguments} &
                & \multicolumn{5}{c}{Non-normative Arguments} \\
            \cmidrule{3-7} \cmidrule{9-13}
             &  & ACC & AUC & F1 & \shortstack{\Fsup} & \shortstack{\Fatt}
                & & ACC & AUC & F1 & \shortstack{\Fsup} & \shortstack{\Fatt} \\
            \midrule
            1 & TGA Net & 66.1 & 75.0 & 65.4 & 69.8 & 60.9 
                & & 66.5 & 74.3 & 65.9 & 70.1 & 61.7 \\
            2 & Hybrid Net & 67.2 & 70.1 & 67.2 & 68.1 & 66.3 
                & & 59.7 & 62.6 & 58.8 & 64.5 & 53.2 \\
            \midrule
            3 & BERT & 79.1 & 88.3 & 79.4 & 79.8 & 79.0 
                & & 80.7 & 87.6 & 80.7 & 81.4 & 79.9  \\
            4 & BERT+LX & 78.4 & 88.1 & 78.4 & 79.2 & 77.5 
                & & \textbf{81.6} & \textbf{88.8} & \textbf{81.5} & \textbf{82.3} & \textbf{80.8} \\
            5 & BERT+MT & 79.6 & 88.2 & 79.6 & 80.0 & 79.1 
                & & 77.6 & 86.3 & 77.5 & 78.9 & 76.0 \\
            \midrule
            6 & LogBERT & \textbf{81.0}$^\star$ & \textbf{88.8} & \textbf{80.7}$^\star$ & \textbf{81.1}$^\star$ & \textbf{80.4}$^\star$   
                & & 81.2 & 88.3 & 80.8 & 81.7 & 80.0 \\ 
            \bottomrule
        \end{tabularx}
        \caption{Debatepedia}
        \label{tab:accs_sup_debate}
    \end{subtable}
    \caption{Accuracy of supervised models. $p$ < \{0.05$^\star$, 0.001$^\ddagger$\} with paired bootstrap compared to BERT.}
    \label{tab:accs_sup}
\end{table*}

\subsection{Baselines}
The first goal of this experiment is to see if the logical mechanisms improve the predictive power of a model trained without concerning them. Thus, our first baseline is \textbf{BERT} fine-tuned on the main task only. This method recently yielded the (near-) best accuracy in argumentative relation classification~\cite{Durmus:2019kialo,reimers-etal-2019-classification}. 

In order to see the effectiveness of the representation learning method, the next two baselines incorporate logical mechanisms in different ways. \textbf{BERT+LX} uses latent cross~\cite{Beutel:2018latent_cross} to directly incorporate predicate values in R1--R13 as features; we use an MLP to encode the predicate values, exploring (i) one hidden layer with $D$=768 and (ii) no hidden layers. BERT+LX consistently outperforms a simple MLP without latent cross. \textbf{BERT+MT} uses multitask learning to train the main and logic tasks simultaneously. 

Lastly, we test two recent models from stance detection and dis/agreement classification.
\textbf{TGA Net} \cite{Allaway:2020stance} takes a statement-topic pair and predicts the statement's stance. It encodes the input using BERT and weighs topic tokens based on similarity to other topics. In our task, claims serve as ``topics''. We use the published implementation, exploring \{50, 100, 150, 200\} for the number of clusters and increasing the max input size to the BERT input size. \textbf{Hybrid Net} \cite{Chen:2018hybrid} takes a quote-response pair and predicts whether the response agrees or disagrees with the quote. It encodes the input using BiLSTM and uses self- and cross-attention between tokens. In our task, claims and statements serve as ``quotes'' and ``responses'', respectively.

\subsection{Results}
Tables \ref{tab:accs_sup_kialo} (Kialo) and \ref{tab:accs_sup_debate} (Debatepedia) summarize the accuracy of each model averaged over 5 runs with random initialization. For non-normative arguments, the causality task is excluded from all models as it consistently hurts them for both datasets. 

Regarding the baselines, TGA Net (row 1) and Hybrid Net (row 2) underperform BERT (row 3). TGA Net, in the original paper, handles topics that are usually short noun phrases. It weighs input topic tokens based on other similar topics, but this method seems not as effective when topics are replaced with longer and more natural claims. Hybrid Net encodes input text using BiLSTM, whose performance is generally inferior to BERT.

BERT trained only on the main task is competitive (row 3). BERT+LX (row 4), which incorporates predicate values directly as features, is comparable to or slightly underperforms BERT in most cases. We speculate that predicate values are not always accurate, so using their values directly can be noisy. 
LogBERT (row 6) consistently outperforms all models except for non-normative arguments in Debatepedia (but it still outperforms BERT).
While both BERT+MT and LogBERT are trained on the same logic tasks, BERT+MT (row 5) performs consistently worse than LogBERT. The reason is likely that logic tasks have much larger training data than the main task, so the model is not optimized enough for the main task. On the other hand, LogBERT is optimized solely for the main task after learning useful representations from the logic tasks, which seem to lay a good foundation for the main task. 

We examined the contribution of each logic task using ablation tests (not shown in the tables). Textual entailment has the strongest contribution across settings, followed by sentiment classification. This contrasts the relatively small contribution of factual consistency in Experiment 1. Moreover, the tasks of normative relation have the smallest contribution for normative arguments and the causality task for non-normative arguments in both datasets. Three of the normative relation tasks take only a statement as input, which is inconsistent with the main task. This inconsistency might cause these tasks to have only small contributions in representation learning. The small contribution of the causality task in both Experiments 1 and 2 suggests large room for improvement in how to effectively operationalize causal relation in argumentation.

\begin{figure}[t]
    \centering
    \begin{subfigure}[t]{\linewidth}
         \centering
         \includegraphics[width=\linewidth]{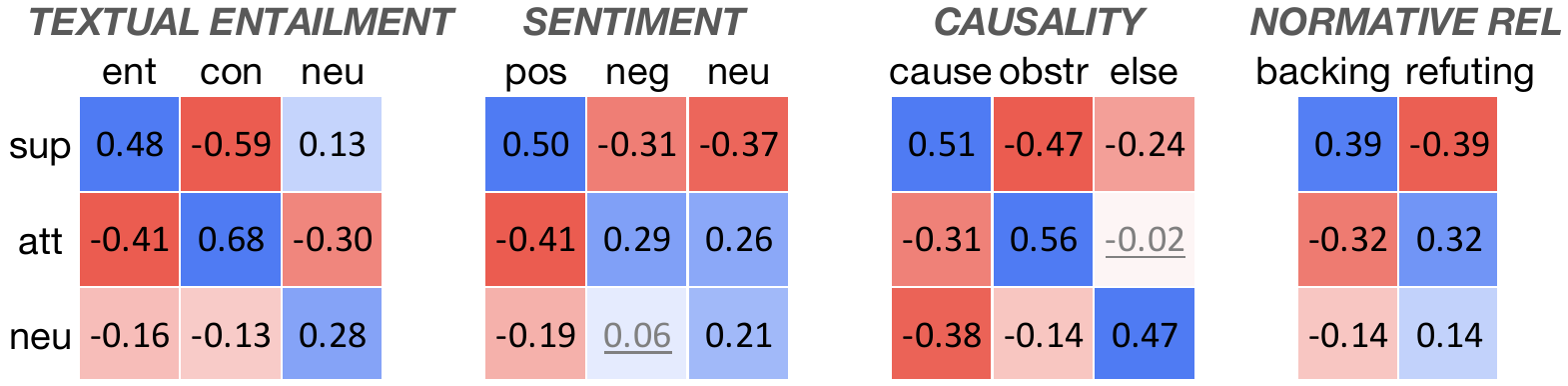}
         \caption{Normative arguments.}
         \label{fig:logbert_corr_normative}
    \end{subfigure}
    \begin{subfigure}[t]{.7\linewidth}
        \centering
        \includegraphics[width=\linewidth]{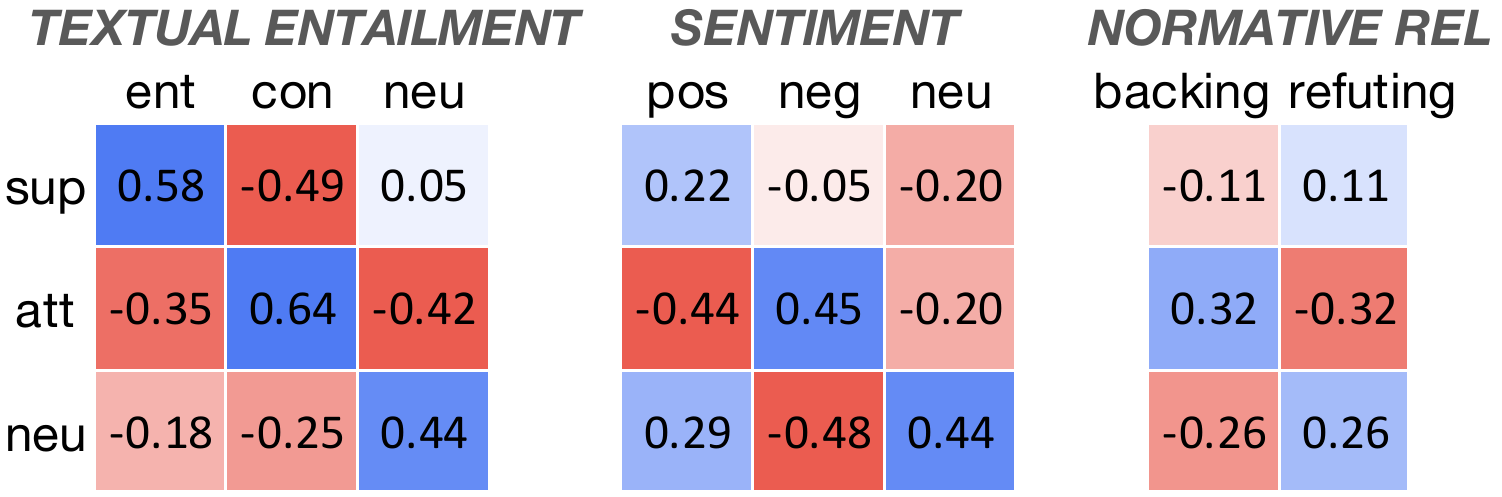}
        \caption{Non-normative arguments.}
        \label{fig:logbert_corr_causal}
    \end{subfigure}
    \caption{Pearson correlation coefficients between argumentative relations and logic tasks from LogBERT. All but underlined values have $p < 0.0001$.}
    \label{fig:logbert_corr}
\end{figure}

To understand how LogBERT makes a connection between the logical relations and argumentative relations, we analyze ``difficult'' arguments in Kialo that BERT misclassified but LogBERT classified correctly. 
If the correct decisions by LogBERT were truly informed by its logic-awareness, the decisions may have correlations with its (internal) decisions for the logic tasks as well, e.g., between attack and textual contradiction. 
Figure \ref{fig:logbert_corr} shows the correlation coefficients between the probabilities of argumentative relations and those of the individual classes of the logic tasks, computed simultaneously by LogBERT (using the pretrained classification layers for the logic tasks). 
For sentiment, the second text of an input pair is the sentiment target, so we can interpret each class roughly as the statement's sentiment toward the claim.
For normative relation, we computed the probabilities of backing (R10+R12) and refuting (R11+R13).

The correlations are intuitive. The support relation is positively correlated with textual entailment, positive sentiment, `cause' of causality, and `backing' of normative relation, whereas the attack relation is positively correlated with textual contradiction, negative sentiment, `obstruct' of causality, and `refuting' of normative relation. The neutral relation is positively correlated with the neutral classes of the logic tasks. The only exception is the normative relation for non-normative arguments. A possible reason is that most claims in non-normative arguments do not follow the typical form of normative claims, and that might affect how the tasks of normative relation contribute for these arguments. 
\hide{We leave a more thorough analysis to future work.}

LogBERT's predictive power comes from its representation of arguments that makes strong correlations between the logical relations and argumentative relations. 
Though LogBERT uses these correlations, it does not necessarily \textit{derive} argumentative relations \textit{from} the logic rules. It is still a black-box model 
with some insightful explainability.

\section{Conclusion}
We examined four types of logical and theory-informed mechanisms in argumentative relations: factual consistency, sentiment coherence, causal relation, and normative relation. 
\hide{We operationalized these mechanisms through machine-learned modules and probabilistic soft logic, to find the optimal argumentative relations between statements.}
To operationalize normative relation, we built rich annotation schema and data for the argumentation schemes \textit{argument from consequences} and \textit{practical reasoning}, too. 

Evaluation on arguments from Kialo and Debatepedia revealed the importance of these mechanisms in argumentation, especially normative relation and sentiment coherence. Their utility was further verified in a supervised setting via our representation learning method. Our model learns argument representations that make strong correlations between logical relations and argumentative relations in intuitive ways. Textual entailment was found to be particularly helpful in the supervised setting.

Some promising future directions are to probe fine-tuned BERT to see if it naturally learns logical mechanisms and to improve PSL with more rules.

\section*{Acknowledgments}
We thank the reviewers and action editor for valuable comments. 
This research was supported in part by the Kwanjeong Educational Foundation, CMU's GuSH Research grant, Volkswagen Stiftung under grant 92 182, ESRC under ES/V003901/1, and EPSRC under EP/N014871/1.

\bibliography{tacl21}

\begin{thebibliography}{51}
\expandafter\ifx\csname natexlab\endcsname\relax\def\natexlab#1{#1}\fi

\bibitem[{Ajjour et~al.(2019)Ajjour, Alshomary, Wachsmuth, and
  Stein}]{Ajjour:2019frame}
Yamen Ajjour, Milad Alshomary, Henning Wachsmuth, and Benno Stein. 2019.
\newblock \href {https://doi.org/10.18653/v1/d19-1290} {{Modeling Frames in
  Argumentation}}.
\newblock In \emph{Proceedings of the 2019 Conference on Empirical Methods in
  Natural Language Processing and the 9th International Joint Conference on
  Natural Language Processing (EMNLP-IJCNLP)}, pages 2922--2932.

\bibitem[{Al-Khatib et~al.(2020)Al-Khatib, Hou, Wachsmuth, Jochim, Bonin, and
  Stein}]{Al-Khatib:2020.causal_graph}
Khalid Al-Khatib, Yufang Hou, Henning Wachsmuth, Charles Jochim, Francesca
  Bonin, and Benno Stein. 2020.
\newblock \href {https://doi.org/10.1609/aaai.v34i05.6231} {{End-to-End
  Argumentation Knowledge Graph Construction}}.
\newblock In \emph{Proceedings of the AAAI Conference on Artificial
  Intelligence}, volume~34, pages 7367--7374.

\bibitem[{Allaway and McKeown(2020)}]{Allaway:2020stance}
Emily Allaway and Kathleen McKeown. 2020.
\newblock \href {https://doi.org/10.18653/v1/2020.emnlp-main.717} {{Zero-Shot
  Stance Detection: A Dataset and Model using Generalized Topic
  Representations}}.
\newblock In \emph{Proceedings of the 2020 Conference on Empirical Methods in
  Natural Language Processing}, pages 8913--8931.

\bibitem[{Bach et~al.(2017)Bach, Broecheler, Huang, and Getoor}]{Bach:2017psl}
Stephen~H. Bach, Matthias Broecheler, Bert Huang, and Lise Getoor. 2017.
\newblock {Hinge-Loss Markov Random Fields and Probabilistic Soft Logic}.
\newblock \emph{Journal of Machine Learning Research}, 18(109):1--67.

\bibitem[{Bakliwal et~al.(2013)Bakliwal, Foster, Puil, Brien, Tounsi, and
  Hughes}]{Bakliwal:irish_senti}
Akshat Bakliwal, Jennifer Foster, Jennifer van~der Puil, Ron~O Brien, Lamia
  Tounsi, and Mark Hughes. 2013.
\newblock {Sentiment Analysis of Political Tweets: Towards an Accurate
  Classifier}.
\newblock In \emph{Proceedings of the Workshop on Language Analysis in Social
  Media}, pages 49--58.

\bibitem[{Beutel et~al.(2018)Beutel, Covington, Jain, Xu, Li, Gatto, and
  Chi}]{Beutel:2018latent_cross}
Alex Beutel, Paul Covington, Sagar Jain, Can Xu, Jia Li, Vince Gatto, and Ed~H.
  Chi. 2018.
\newblock \href {https://doi.org/10.1145/3159652.3159727} {{Latent Cross:
  Making Use of Context in Recurrent Recommender Systems}}.
\newblock In \emph{Proceedings of the Eleventh ACM International Conference on
  Web Search and Data Mining}, page 46–54, New York, NY, USA. Association for
  Computing Machinery.

\bibitem[{Cabrio et~al.(2013)Cabrio, Tonelli, and
  Villata}]{Cabrio.2013.schemes}
Elena Cabrio, Sara Tonelli, and Serena Villata. 2013.
\newblock {From Discourse Analysis to Argumentation Schemes and Back: Relations
  and Differences}.
\newblock In \emph{Computational Logic in Multi-Agent Systems}, pages 1--17,
  Berlin, Heidelberg. Springer Berlin Heidelberg.

\bibitem[{Cabrio and Villata(2012)}]{Cabrio:2012ud}
Elena Cabrio and Serena Villata. 2012.
\newblock {Combining Textual Entailment and Argumentation Theory for Supporting
  Online Debates Interactions}.
\newblock In \emph{Proceedings of the 50th Annual Meeting of the Association
  for Computational Linguistics (Volume 2: Short Papers)}, pages 208--212.
  Association for Computational Linguistics.

\bibitem[{Chakrabarty et~al.(2019)Chakrabarty, Hidey, Muresan, McKeown, and
  Hwang}]{Chakrabarty:2019ampersand}
Tuhin Chakrabarty, Christopher Hidey, Smaranda Muresan, Kathy McKeown, and
  Alyssa Hwang. 2019.
\newblock \href {https://doi.org/10.18653/v1/d19-1291} {{AMPERSAND: Argument
  Mining for PERSuAsive oNline Discussions}}.
\newblock In \emph{Proceedings of the 2019 Conference on Empirical Methods in
  Natural Language Processing and the 9th International Joint Conference on
  Natural Language Processing (EMNLP-IJCNLP)}, pages 2926--2936.

\bibitem[{Chen et~al.(2018)Chen, Du, Bing, and Xu}]{Chen:2018hybrid}
Di~Chen, Jiachen Du, Lidong Bing, and Ruifeng Xu. 2018.
\newblock {Hybrid Neural Attention for Agreement/Disagreement Inference in
  Online Debates}.
\newblock In \emph{Proceedings of the 2018 Conference of Empirical Methods in
  Natural Language Processing}, pages 665--670.

\bibitem[{Choi and Lee(2018)}]{choi-lee-2018-gist}
HongSeok Choi and Hyunju Lee. 2018.
\newblock \href {https://doi.org/10.18653/v1/s18-1122} {{GIST at SemEval-2018
  Task 12: A network transferring inference knowledge to Argument Reasoning
  Comprehension task}}.
\newblock In \emph{Proceedings of The 12th International Workshop on Semantic
  Evaluation}, pages 773--777.

\bibitem[{Devlin et~al.(2019)Devlin, Chang, Lee, and
  Toutanova}]{Devlin:2018bert}
Jacob Devlin, Ming-Wei Chang, Kenton Lee, and Kristina Toutanova. 2019.
\newblock {BERT: Pre-training of Deep Bidirectional Transformers for Language
  Understanding}.
\newblock In \emph{Proceedings of the 2019 Conference of the North American
  Chapter of the Association for Computational Linguistics: Human Language
  Technologies, Volume 1 (Long and Short Papers)}, pages 4171--4186.

\bibitem[{Dong et~al.(2014)Dong, Wei, Tan, Tang, Zhou, and
  Xu}]{dong-etal-2014-adaptive}
Li~Dong, Furu Wei, Chuanqi Tan, Duyu Tang, Ming Zhou, and Ke~Xu. 2014.
\newblock \href {https://doi.org/10.3115/v1/p14-2009} {{Adaptive Recursive
  Neural Network for Target-dependent Twitter Sentiment Classification}}.
\newblock In \emph{Proceedings of the 52nd Annual Meeting of the Association
  for Computational Linguistics (Volume 2: Short Papers)}, pages 49--54.

\bibitem[{Dunietz et~al.(2017)Dunietz, Levin, and
  Carbonell}]{Dunietz17:because}
Jesse Dunietz, Lori Levin, and Jaime Carbonell. 2017.
\newblock \href {https://doi.org/10.18653/v1/w17-0812} {{The BECauSE Corpus
  2.0: Annotating Causality and Overlapping Relations}}.
\newblock In \emph{Proceedings of the 11th Linguistic Annotation Workshop},
  pages 95--104.

\bibitem[{Durmus et~al.(2019)Durmus, Ladhak, and Cardie}]{Durmus:2019kialo}
Esin Durmus, Faisal Ladhak, and Claire Cardie. 2019.
\newblock {Determining Relative Argument Specificity and Stance for Complex
  Argumentative Structures}.
\newblock In \emph{Proceedings of the 57th Annual Meeting of the Association
  for Computational Linguistics}, pages 4630--4641.

\bibitem[{Eger et~al.(2017)Eger, Daxenberger, and Gurevych}]{Eger:2017e2e}
Steffen Eger, Johannes Daxenberger, and Iryna Gurevych. 2017.
\newblock \href {https://doi.org/10.18653/v1/p17-1002} {{Neural End-to-End
  Learning for Computational Argumentation Mining}}.
\newblock In \emph{Proceedings of the 55th Annual Meeting of the Association
  for Computational Linguistics (Volume 1: Long Papers)}, pages 11--22.

\bibitem[{Feng and Hirst(2011)}]{Feng:2011vp}
Vanessa~Wei Feng and Graeme Hirst. 2011.
\newblock {Classifying arguments by scheme}.
\newblock In \emph{Proceedings of the 49th Annual Meeting of the Association
  for Computational Linguistics: Human Language Technologies}, pages 987--996,
  Portland, Oregon, USA. Association for Computational Linguistics.

\bibitem[{Gemechu and Reed(2019)}]{Gemechu2020:acl}
Debela Gemechu and Chris Reed. 2019.
\newblock {Decompositional Argument Mining: A General Purpose Approach for
  Argument Graph Construction}.
\newblock In \emph{Proceedings of the 57th Annual Meeting of the Association
  for Computational Linguistics}, pages 516--526.

\bibitem[{Go et~al.(2009)Go, Bhayani, and Huang}]{Go:2009senti140}
Alec Go, Richa Bhayani, and Lei Huang. 2009.
\newblock {Twitter sentiment classification using distant supervision}.
\newblock \emph{CS224N Project Report}.

\bibitem[{Habernal and Gurevych(2017)}]{Habernal:2017cl}
Ivan Habernal and Iryna Gurevych. 2017.
\newblock \href {https://doi.org/10.1162/coli\_a\_00276} {{Argumentation Mining
  in User-Generated Web Discourse}}.
\newblock \emph{Computational Linguistics}, 43(1):125--179.

\bibitem[{Hou and Jochim(2017)}]{Hou.2017}
Yufang Hou and Charles Jochim. 2017.
\newblock \href {https://doi.org/10.18653/v1/w17-5107} {{Argument Relation
  Classification Using a Joint Inference Model}}.
\newblock In \emph{Proceedings of the 4th Workshop on Argument Mining}, pages
  60--66.

\bibitem[{Jo et~al.(2020)Jo, Mayfield, Reed, and Hovy}]{Jo:2020lrec}
Yohan Jo, Elijah Mayfield, Chris Reed, and Eduard Hovy. 2020.
\newblock {Machine-Aided Annotation for Fine-Grained Proposition Types in
  Argumentation}.
\newblock In \emph{Proceedings of the 12th International Conference on Language
  Resources and Evaluation}, pages 1008--1018.

\bibitem[{Kobbe et~al.(2020)Kobbe, Hulpu{\textcommabelow{s}}, and
  Stuckenschmidt}]{Kobbe.2020}
Jonathan Kobbe, Ioana Hulpu{\textcommabelow{s}}, and Heiner Stuckenschmidt.
  2020.
\newblock \href {https://doi.org/10.18653/v1/2020.emnlp-main.4} {{Unsupervised
  stance detection for arguments from consequences}}.
\newblock In \emph{Proceedings of the 2020 Conference on Empirical Methods in
  Natural Language Processing (EMNLP)}, pages 50--60, Online. Association for
  Computational Linguistics.

\bibitem[{Lawrence and Reed(2016)}]{Lawrence:2016kt}
John Lawrence and Chris Reed. 2016.
\newblock {Argument Mining Using Argumentation Scheme Structures}.
\newblock In \emph{Proceedings of the Sixth International Conference on
  Computational Models of Argument}, pages 379--390.

\bibitem[{Lawrence and Reed(2017)}]{Lawrence:2017topic}
John Lawrence and Chris Reed. 2017.
\newblock \href {https://doi.org/10.18653/v1/w17-5105} {{Mining Argumentative
  Structure from Natural Language text using Automatically Generated
  Premise-Conclusion Topic Models}}.
\newblock In \emph{Proceedings of the 4th Workshop on Argument Mining}, pages
  39--48.

\bibitem[{Lawrence et~al.(2019)Lawrence, Visser, and Reed}]{Lawrence:2019we}
John Lawrence, Jacky Visser, and Chris Reed. 2019.
\newblock {An Online Annotation Assistant for Argument Schemes}.
\newblock In \emph{Proceedings of the 13th Linguistic Annotation Workshop},
  pages 100--107.

\bibitem[{Lindahl et~al.(2019)Lindahl, Borin, and Rouces}]{Lindahl:2019uc}
Anna Lindahl, Lars Borin, and Jacobo Rouces. 2019.
\newblock {Towards Assessing Argumentation Annotation - A First Step}.
\newblock In \emph{Proceedings of the 6th Workshop on Argument Mining}, pages
  177--186.

\bibitem[{Mitchell et~al.(2013)Mitchell, Aguilar, Wilson, and
  Durme}]{mitchell-etal-2013-open}
Margaret Mitchell, Jacqui Aguilar, Theresa Wilson, and Benjamin~Van Durme.
  2013.
\newblock {Open Domain Targeted Sentiment}.
\newblock In \emph{Proceedings of the 2013 Conference on Empirical Methods in
  Natural Language Processing}, pages 1643--1654.

\bibitem[{Nguyen et~al.(2017)Nguyen, Walde, and Vu}]{Nguyen.2017antonyms}
Kim~Anh Nguyen, Sabine Schulte~im Walde, and Ngoc~Thang Vu. 2017.
\newblock {Distinguishing Antonyms and Synonyms in a Pattern-based Neural
  Network}.
\newblock In \emph{Proceedings of the 15th Conference of the European Chapter
  of the Association for Computational Linguistics: Volume 1, Long Papers},
  pages 76--85, Valencia, Spain. Association for Computational Linguistics.

\bibitem[{Niven and Kao(2019)}]{Niven:2019cq}
Timothy Niven and Hung-Yu Kao. 2019.
\newblock \href {https://doi.org/10.18653/v1/p19-1459} {{Probing Neural Network
  Comprehension of Natural Language Arguments}}.
\newblock In \emph{Proceedings of the 57th Annual Meeting of the Association
  for Computational Linguistics}, pages 4658--4664.

\bibitem[{Opitz and Frank(2019)}]{Opitz:2019tk}
Juri Opitz and Anette Frank. 2019.
\newblock {Dissecting Content and Context in Argumentative Relation Analysis}.
\newblock In \emph{Proceedings of the 6th Workshop on Argument Mining}, pages
  25--34.

\bibitem[{Park and Cardie(2018)}]{Park:2018wy}
Joonsuk Park and Claire Cardie. 2018.
\newblock {A Corpus of eRulemaking User Comments for Measuring Evaluability of
  Arguments}.
\newblock In \emph{Proceedings of the Eleventh International Conference on
  Language Resources and Evaluation (LREC-2018)}.

\bibitem[{Persing and Ng(2016)}]{Persing:2016e2e}
Isaac Persing and Vincent Ng. 2016.
\newblock \href {https://doi.org/10.18653/v1/n16-1164} {{End-to-End
  Argumentation Mining in Student Essays}}.
\newblock In \emph{Proceedings of the 2016 Conference of the North American
  Chapter of the Association for Computational Linguistics: Human Language
  Technologies}, pages 1384--1394.

\bibitem[{Reed et~al.(2008)Reed, Palau, Rowe, and Moens}]{Reed.2008.araucaria}
Chris Reed, Raquel~Mochales Palau, Glenn Rowe, and Marie-Francine Moens. 2008.
\newblock \href
  {http://www.lrec-conf.org/proceedings/lrec2008/pdf/648\_paper.pdf} {{Language
  Resources for Studying Argument}}.
\newblock In \emph{Proceedings of the Sixth International Conference on
  Language Resources and Evaluation ({LREC}'08)}, Marrakech, Morocco. European
  Language Resources Association (ELRA).

\bibitem[{Reimers et~al.(2019)Reimers, Schiller, Beck, Daxenberger, Stab, and
  Gurevych}]{reimers-etal-2019-classification}
Nils Reimers, Benjamin Schiller, Tilman Beck, Johannes Daxenberger, Christian
  Stab, and Iryna Gurevych. 2019.
\newblock \href {https://doi.org/10.18653/v1/p19-1054} {{Classification and
  Clustering of Arguments with Contextualized Word Embeddings}}.
\newblock In \emph{Proceedings of the 57th Annual Meeting of the Association
  for Computational Linguistics}, pages 567--578.

\bibitem[{Reisert et~al.(2018)Reisert, Inoue, Kuribayashi, and
  Inui}]{Reisert.2018.schemes}
Paul Reisert, Naoya Inoue, Tatsuki Kuribayashi, and Kentaro Inui. 2018.
\newblock \href {https://doi.org/10.18653/v1/w18-5210} {{Feasible Annotation
  Scheme for Capturing Policy Argument Reasoning using Argument Templates}}.
\newblock In \emph{Proceedings of the 5th Workshop on Argument Mining}, pages
  79--89, Brussels, Belgium. Association for Computational Linguistics.

\bibitem[{Rinott et~al.(2015)Rinott, Dankin, Perez, Khapra, Aharoni, and
  Slonim}]{Rinott.2015.evidence}
Ruty Rinott, Lena Dankin, Carlos~Alzate Perez, Mitesh~M. Khapra, Ehud Aharoni,
  and Noam Slonim. 2015.
\newblock \href {https://doi.org/10.18653/v1/d15-1050} {{Show Me Your Evidence
  - an Automatic Method for Context Dependent Evidence Detection}}.
\newblock In \emph{Proceedings of the 2015 Conference on Empirical Methods in
  Natural Language Processing}, pages 440--450, Lisbon, Portugal. Association
  for Computational Linguistics.

\bibitem[{Rosenthal et~al.(2017)Rosenthal, Farra, and Nakov}]{semeval17task4}
Sara Rosenthal, Noura Farra, and Preslav Nakov. 2017.
\newblock \href {https://doi.org/10.18653/v1/s17-2088} {{SemEval-2017 Task 4:
  Sentiment Analysis in Twitter}}.
\newblock In \emph{Proceedings of the 11th International Workshop on Semantic
  Evaluation (SemEval-2017)}, pages 502--518.

\bibitem[{Rosenthal and McKeown(2015)}]{Rosenthal:2015iw}
Sara Rosenthal and Kathy McKeown. 2015.
\newblock {I Couldn't Agree More: The Role of Conversational Structure in
  Agreement and Disagreement Detection in Online Discussions}.
\newblock In \emph{Proceedings of the 16th Annual Meeting of the Special
  Interest Group on Discourse and Dialogue}, pages 168--177. Association for
  Computational Linguistics.

\bibitem[{Speer et~al.(2017)Speer, Chin, and Havasi}]{Speer17:conceptnet}
Robyn Speer, Joshua Chin, and Catherine Havasi. 2017.
\newblock \href {https://aaai.org/ocs/index.php/AAAI/AAAI17/paper/view/14972}
  {{ConceptNet 5.5: An Open Multilingual Graph of General Knowledge}}.
\newblock In \emph{Proceedings of the Thirty-First AAAI Conference on
  Artificial Intelligence}.

\bibitem[{Stab and Gurevych(2017)}]{Stab:2017cl}
Christian Stab and Iryna Gurevych. 2017.
\newblock \href {https://doi.org/10.1162/coli\_a\_00295} {{Parsing
  Argumentation Structures in Persuasive Essays}}.
\newblock \emph{Computational Linguistics}, 1(1):1--62.

\bibitem[{Stab et~al.(2018)Stab, Miller, and Gurevych}]{Stab:2018cross-topic}
Christian Stab, Tristan Miller, and Iryna Gurevych. 2018.
\newblock {Cross-topic Argument Mining from Heterogeneous Sources Using
  Attention-based Neural Networks}.
\newblock In \emph{Proceedings of the 2018 Conference on Empirical Methods in
  Natural Language Processing}, pages 3664--3674.

\bibitem[{Tandon et~al.(2019)Tandon, Mishra, Sakaguchi, Bosselut, and
  Clark}]{Tandon19:wiqa}
Niket Tandon, Bhavana~Dalvi Mishra, Keisuke Sakaguchi, Antoine Bosselut, and
  Peter Clark. 2019.
\newblock \href {https://doi.org/10.18653/v1/d19-1629} {{WIQA: A dataset for
  ``What if...'' reasoning over procedural text}}.
\newblock In \emph{Proceedings of the 2019 Conference on Empirical Methods in
  Natural Language Processing and the 9th International Joint Conference on
  Natural Language Processing (EMNLP-IJCNLP)}, pages 6076--6085.

\bibitem[{Visser et~al.(2020)Visser, Lawrence, Reed, Wagemans, and
  Walton}]{Visser.2020.Argumentation}
Jacky Visser, John Lawrence, Chris Reed, Jean Wagemans, and Douglas Walton.
  2020.
\newblock \href {https://doi.org/10.1007/s10503-020-09519-x} {{Annotating
  Argument Schemes}}.
\newblock \emph{Argumentation}, pages 1--39.

\bibitem[{Walton et~al.(2008)Walton, Reed, and Macagno}]{Walton:2008schem}
Douglas Walton, Chris Reed, and Fabrizio Macagno. 2008.
\newblock \emph{{Argumentation Schemes}}.
\newblock Cambridge University Press.

\bibitem[{Webber et~al.(2006)Webber, Prasad, Lee, and Joshi}]{Webber:2006pdtb3}
Bonnie Webber, Rashmi Prasad, Alan Lee, and Aravind Joshi. 2006.
\newblock {The Penn Discourse Treebank 3.0 Annotation Manual}.

\bibitem[{Williams et~al.(2018)Williams, Nangia, and
  Bowman}]{Williams:2018mnli}
Adina Williams, Nikita Nangia, and Samuel Bowman. 2018.
\newblock \href {https://doi.org/10.18653/v1/n18-1101} {{A Broad-Coverage
  Challenge Corpus for Sentence Understanding through Inference}}.
\newblock In \emph{Proceedings of the 2018 Conference of the North American
  Chapter of the Association for Computational Linguistics: Human Language
  Technologies, Volume 1 (Long Papers)}, pages 1112--1122.

\bibitem[{Wilson et~al.(2005)Wilson, Wiebe, and Hoffmann}]{Wilson:2005subjlex}
Theresa Wilson, Janyce Wiebe, and Paul Hoffmann. 2005.
\newblock {Recognizing Contextual Polarity in Phrase-Level Sentiment Analysis}.
\newblock In \emph{Proceedings of Human Language Technology Conference and
  Conference on Empirical Methods in Natural Language Processing}, pages
  347--354, Vancouver, British Columbia, Canada. Association for Computational
  Linguistics.

\bibitem[{Wolf et~al.(2020)Wolf, Debut, Sanh, Chaumond, Delangue, Moi, Cistac,
  Rault, Louf, Funtowicz, Davison, Shleifer, Platen, Ma, Jernite, Plu, Xu,
  Scao, Gugger, Drame, Lhoest, and Rush}]{huggingface2020}
Thomas Wolf, Lysandre Debut, Victor Sanh, Julien Chaumond, Clement Delangue,
  Anthony Moi, Pierric Cistac, Tim Rault, Rémi Louf, Morgan Funtowicz, Joe
  Davison, Sam Shleifer, Patrick~von Platen, Clara Ma, Yacine Jernite, Julien
  Plu, Canwen Xu, Teven~Le Scao, Sylvain Gugger, Mariama Drame, Quentin Lhoest,
  and Alexander~M. Rush. 2020.
\newblock \href {https://doi.org/10.18653/v1/2020.emnlp-demos.6}
  {{Transformers: State-of-the-art Natural Language Processing}}.
\newblock In \emph{Proceedings of the 2020 Conference on Empirical Methods in
  Natural Language Processing: System Demonstrations}, pages 38--45.
  Association for Computational Linguistics.

\bibitem[{Xu et~al.(2018)Xu, Paris, Nepal, and Sparks}]{Xu.2018}
Chang Xu, C\'ecile Paris, Surya Nepal, and Ross Sparks. 2018.
\newblock \href {https://doi.org/10.18653/v1/p18-2123} {{Cross-Target Stance
  Classification with Self-Attention Networks}}.
\newblock In \emph{Proceedings of the 56th Annual Meeting of the Association
  for Computational Linguistics (Volume 2: Short Papers)}, pages 778--783,
  Melbourne, Australia. Association for Computational Linguistics.

\bibitem[{Xu et~al.(2019)Xu, Paris, Nepal, and Sparks}]{Xu:2019reason}
Chang Xu, Cecile Paris, Surya Nepal, and Ross Sparks. 2019.
\newblock \href {https://doi.org/10.18653/v1/p19-1460} {{Recognising Agreement
  and Disagreement between Stances with Reason Comparing Networks}}.
\newblock In \emph{Proceedings of the 57th Annual Meeting of the Association
  for Computational Linguistics}, pages 4665--4671.

\end{thebibliography}
\bibliographystyle{acl_natbib}

\end{document}